\documentclass[10pt,twocolumn,letterpaper]{article}

\usepackage{cvpr}              

\usepackage{graphicx}
\makeatletter
\@namedef{ver@everyshi.sty}{}
\makeatother
\usepackage{amsmath}
\usepackage{amssymb}
\usepackage{booktabs}
\usepackage{stfloats}
\usepackage{nicematrix}
\usepackage{soul}
\usepackage[accsupp]{axessibility}
\usepackage{color}
\usepackage{makecell}
\usepackage[shortlabels]{enumitem}

\usepackage[pagebackref,breaklinks,colorlinks]{hyperref}
\usepackage[capitalize]{cleveref}
\usepackage{enumitem}

\crefname{section}{Sec.}{Secs.}
\Crefname{section}{Section}{Sections}
\Crefname{table}{Table}{Tables}
\crefname{table}{Tab.}{Tabs.}
\usepackage{caption}
\renewcommand{\raggedright}{\leftskip=0pt \rightskip=0pt plus 0cm}
\usepackage{array}
\usepackage{changepage}
\usepackage{multirow}

\usepackage{pifont}
\def\cvprPaperID{6496} 
\def\confName{CVPR}
\def\confYear{2023}

\newcommand{\Tref}[1]{Table~\ref{#1}}
\newcommand{\Eref}[1]{Eq.~(\ref{#1})}
\newcommand{\Fref}[1]{Fig.~\ref{#1}}
\newcommand{\Sref}[1]{Sec.~\ref{#1}}

\newcommand{\eref}[1]{Eq.~(\ref{#1})}

\begin{document}

\title{DANI-Net: Uncalibrated Photometric Stereo by Differentiable Shadow Handling, Anisotropic Reflectance Modeling, and Neural Inverse Rendering}


\author{
Zongrui~Li$^1$~~~
Qian Zheng$^{2, 3,}$\thanks{Corresponding author}~~~
Boxin~Shi$^{4, 5}$~~~
Gang~Pan$^{2, 3}$~~~
Xudong Jiang$^1$~~~
\smallskip 
\\
$^1${\small School of Electrical and Electronic Engineering, Nanyang Technological University, Singapore}\\
$^2${\small The State Key Lab of Brain-Machine Intelligence, Zhejiang University, Hangzhou, China}\\
$^3${\small College of Computer Science and Technology, Zhejiang University, Hangzhou, China}\\
$^4${\small National Key Laboratory for Multimedia Information Processing, School of Computer Science, Peking University, Beijing, China}\\
$^5${\small National Engineering Research Center of Visual Technology, School of Computer Science, Peking University, Beijing, China}\\
{\tt\small \{zongrui001, EXDJiang\}@ntu.edu.sg},~~
{\tt\small \{qianzheng,gpan\}@zju.edu.cn},~~
{\tt\small shiboxin@pku.edu.cn}
}

\maketitle

\begin{abstract}
Uncalibrated photometric stereo (UPS) is challenging due to the inherent ambiguity brought by the unknown light. Although the ambiguity is alleviated on non-Lambertian objects, the problem is still difficult to solve for more general objects with complex shapes introducing irregular shadows and general materials with complex reflectance like anisotropic reflectance. To exploit cues from shadow and reflectance to solve UPS and improve performance on general materials, we propose DANI-Net, an inverse rendering framework with differentiable shadow handling and anisotropic reflectance modeling. 
Unlike most previous methods that use non-differentiable shadow maps and assume isotropic material, our network benefits from cues of shadow and anisotropic reflectance through two differentiable paths. 
Experiments on multiple real-world datasets demonstrate our superior and robust performance.
\end{abstract}

\section{Introduction}
\label{sec:intro}
Photometric stereo (PS)~\cite{woodham1980} aims at recovering the surface normal from several images captured under varying light conditions with a fixed viewpoint. 
It has been applied to many fields (\eg, movies production~\cite{chabert2006relighting}, industrial quality inspection \cite{weigl2015photometric}, and biometrics \cite{xie2013real}) due to its advantage in recovering fine-detailed surfaces over other approaches~\cite{herbort2011introduction,chen2020learned} (\eg, multi-view stereo~\cite{seitz2006comparison}, active sensor-based solutions~\cite{zhang2002novel}).
Light calibration is crucial to the performance~\cite{xie2015practical}. However, it is also tedious, restricting the applicability of PS. Therefore, uncalibrated photometric stereo (UPS) methods estimating surface normal with unknown lights have been widely studied in the literature.

Uncalibrated photometric stereo suffers from General Bas-Relief (GBR) ambiguity~\cite{belhumeur1999} for an integrable, Lambertian surface. However, GBR ambiguity is alleviated on a non-Lambertian surface~\cite{georghiades2003incorporating}.
Therefore, recent advances in UPS (\eg,~\cite{li2022self, yang2022ps}) adopt the isotropic reflectance model accounting for non-Lambertian effects to solve UPS. Nonetheless, such a model restricts methods' performance on objects with more general (\eg, anisotropic) materials, while modeling general reflectance is challenging due to extra unknowns, which eventually make UPS intractable. 
Other works (\eg,~\cite{li2022neural, yang2022s, yang2022ps}) notice the benefits of the shadow cues in utilizing global shape-light information to solve PS/UPS because the shadow reflects the interaction of shape and light~\cite{yu2002shadow, kriegman2001shadows}.
However, these methods either fail to exploit the shadow cues due to the lack of a differentiable path from the shadow to the concerned unknowns like shape~\cite{li2022neural}, or the shadow cues have limited effects on the visible shape reconstruction due to the implicit shape representation~\cite{yang2022s, yang2022ps}. 

\begin{figure}[t]
    \centering
    \includegraphics[width=0.47 \textwidth]{"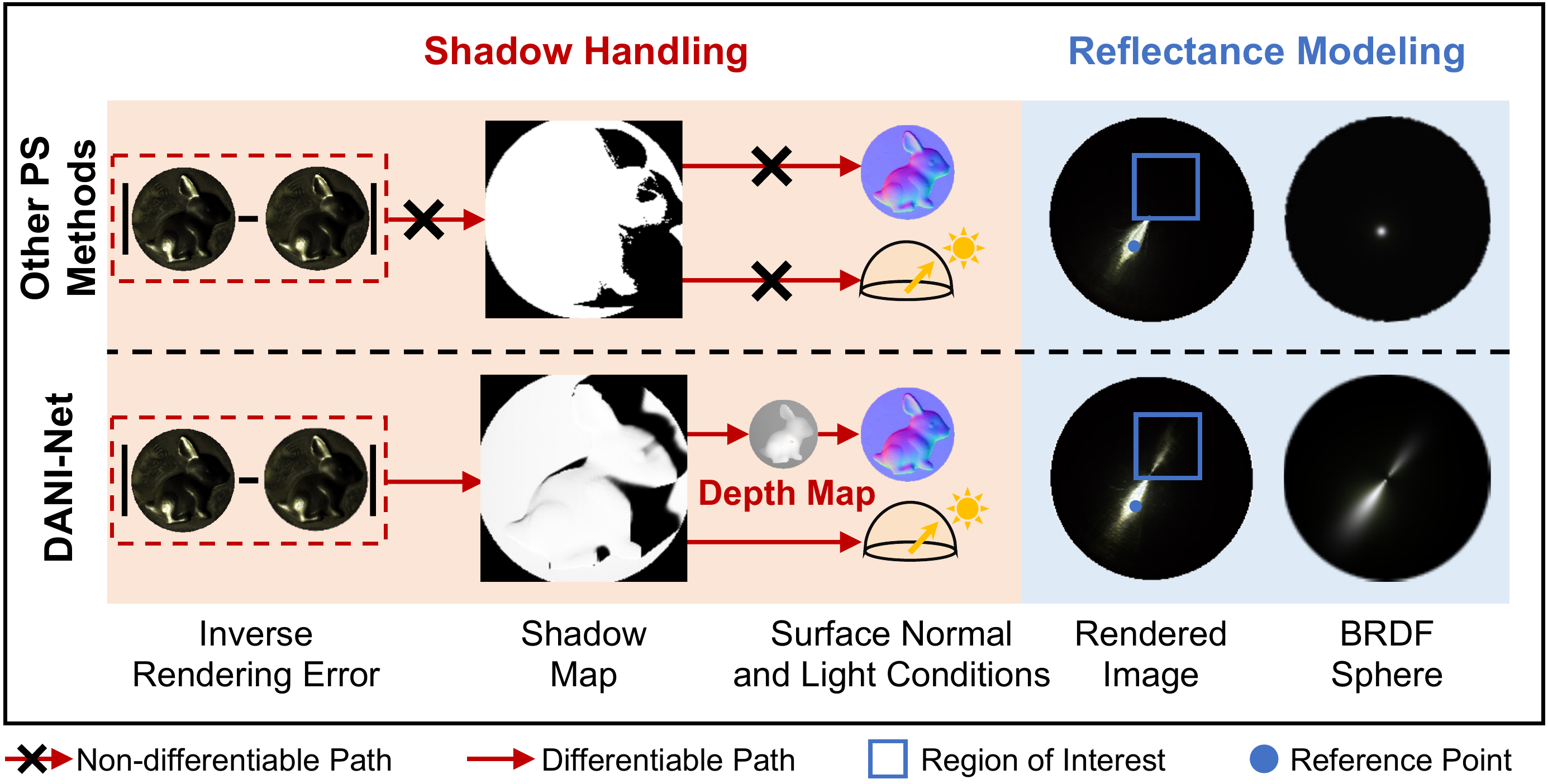"} 
    \vspace{-5pt}
    \caption{
    The proposed DANI-Net differs from other PS or UPS methods in two aspects: 1) {\bf Shadow Handling.} The path in the sequence of inverse rendering errors, shadow maps, and surface normal maps (or light conditions) of the DANI-Net is differentiable; 2) {\bf Reflectance Modeling.} DANI-Net adopts an anisotropic reflectance model. 
    The state-of-the-art UPS method SCPS-NIR~\cite{li2022self} is compared in this figure.
    As can be observed, the proposed DANI-Net produces a smoother and more realistic shadow map of copper {\sc Bunny} thanks to the differentiable shadow handling and renders a more realistic copper {\sc Ball} image and Bidirectional Reflectance Distribution Function (BRDF)  sphere (of the reference point) due to the anisotropic reflectance modeling.
    Data of copper {\sc Bunny} and {\sc Ball} are from {\sc DiLiGenT$10^2$}~\cite{ren2022diligent102}. 
    } 
    \label{fig:teaser}
    \vspace{-10pt}
\end{figure}

To this end, this paper proposes the {\bf DANI-Net}, which solves UPS by {\bf D}ifferentiable shadow handling, {\bf A}nisotropic reflectance modeling, and {\bf N}eural {\bf I}nverse Rendering. 
DANI-Net builds the differentiable path in the sequence of inverse rendering errors, shadow maps, and surface normal maps (or light conditions) (\Fref{fig:teaser}) to fully exploit the shadow cues to solve UPS.
Since those cues facilitate solving extra unknowns introduced by a more sophisticated reflectance model, DANI-Net manages to build up such a model (\Fref{fig:teaser}) to improve the performance on general materials.
During optimization, DANI-Net propagates inverse rendering errors via two paths of shadow cues and anisotropic reflectance, respectively, and simultaneously optimizes the shape (\ie, the depth map and surface normal map), anisotropic reflectance model, shadow map, and light conditions (\ie, direction and intensity). As a result, DANI-Net achieves state-of-the-art performance on several real-world benchmark datasets.
In a nutshell, our contributions are summarized as follows:
\begin{itemize}[itemsep=2.pt,topsep=2.pt,parsep=2.pt]
    \item We propose a differentiable shadow handling method that facilitates exploiting shadow cues with global shape-light information to solve UPS. Experimental results demonstrate its effectiveness in shadow map recovery and surface normal estimation.
    \item We introduce an anisotropic reflectance model that describes both isotropic and anisotropic materials to improve performance on general materials. Experimental results demonstrate its effectiveness on surface normal estimation for objects with a broad range of isotropic and anisotropic materials.
    \item We propose the DANI-Net that simultaneously optimizes shape, anisotropic reflectance, shadow map, and light conditions in an unsupervised manner, propagating inverse rendering errors through two paths involving the shadow cues and anisotropic reflectance, respectively.
    DANI-Net achieves state-of-the-art performance on several real-world benchmark datasets.
\end{itemize}

\section{Related Work}

This section reviews the relevant works in PS. 
\Tref{tab:method} summarizes differences between representative existing PS methods and the proposed DANI-Net. We also briefly review recent advances in neural reflectance representation in 3D vision. 
Note that our reviews on calibrated PS only include the unsupervised calibrated PS methods. Readers may refer to~\cite{shi2016benchmark,xie2022neural,zheng2020summary} for more summaries on supervised PS methods and neural reflectance representation methods. 

\textbf{Unsupervised calibrated photometric stereo.}
The baseline method (LS~\cite{woodham1980}) assumes Lambertian surface and solves PS via least squares optimization.
A category of traditional methods considers the non-Lambertian reflectance as outliers~\cite{barsky20034, chung2008efficient, wu2010robust, wu2006dense}. 
Another category of traditional methods either adopt analytic models (\eg, Torrance-Sparrow~\cite{georghiades2003incorporating}, Ward~\cite{chung2008efficient,goldman2009shape,ackermann2012photometric}, Bi-polynomial~\cite{zheng2019numerical}, etc.) or utilize the general reflectance features (\eg, isotropy~\cite{alldrin2008photometric}, monotonicity~\cite{higo2010consensus}, anisotropic properties~\cite{goldman2009shape, holroyd2008photometric}). 
Traditional methods rely on optimizers tailored to specific assumptions, making them computationally efficient but less accurate. In contrast, although learning-based approaches are more computationally demanding, they could offer superior performance on general objects.
Recently, a group of learning-based unsupervised methods (TM18~\cite{taniai2018neural} and LL22~\cite{li2022neural}) has been proposed to estimate the spatially varying BRDFs and the surface normal with known lights.

\textbf{Uncalibrated photometric stereo.}
Early works either hold the Lambertian assumption and exploit extra clues from reflectance~\cite{lu2017symps,shi2010self,yeung2014normal} or make additional assumptions of light source distribution~\cite{zhou2010ring, shi2010self, wu2013calibrating, papadhimitri2014closed, lu2017symps} to alleviate GBR ambiguity~\cite{belhumeur1999} in UPS. 
Supervised methods (CH19~\cite{chen2019self}, CW20~\cite{chen2020learned}, and SK22~\cite{sarno2022neural}) achieve promising performance on public benchmark datasets.
However, these methods assume the light intensity distributing in a pre-defined range (\ie, $[0.2, 2]$) and solve UPS in two-stage, making them suffer from the data bias (between synthetic training data and real-world ones) and the accumulating errors.
Recently, SCPS-NIR~\cite{li2022self} utilizes the neural inverse rendering method to jointly optimize light and surface normal in an unsupervised manner based on local reflectance information, free from data bias and accumulating errors.
The proposed DANI-Net also addresses UPS in an unsupervised manner, but it differs from all previous works in two aspects. 
1) Our method builds a differentiable path in the sequence of inverse rendering errors, shadow maps, and surface normal maps (or light conditions), which fully exploits shadow cues containing global shape-light information to solve UPS.
2) Our method introduces the anisotropic reflectance model in solving UPS, which improves the performance on general materials.
Besides, as compared with~\cite{li2022self} that calculates shadow maps by image binarization and fixes them during training, our method computes shadow maps from shapes and constantly updates them during training.

\textbf{Shadow handling in photometric stereo.} 
Supervised learning-based methods~\cite{chen2019self, chen2020learned, sarno2022neural} handle shadow implicitly by learning priors from training data, while unsupervised ones often consider the shadow as the outliers (\eg,~\cite{woodham1980,taniai2018neural,papadhimitri2014closed,yuille1997shape,li2022neural,li2022self}).
Aware of useful cues in the shadow, several works utilize the shadow information explicitly by involving the shadow in inverse rendering to solve PS/UPS. 
However, early works~\cite{chandraker2007shadowcuts, sunkavalli2010visibility} assume Lambertian surface, which restricts them to general materials, while
recent advances (LL22~\cite{li2022neural}, SCPS-NIR~\cite{li2022self}, S3-NeRF~\cite{yang2022s}, KF22~\cite{karnieli2022deepshadow}, PS-NeRF~\cite{yang2022ps}) fail to efficiently exploit shadow cues.
That is, some works~\cite{li2022self, li2022neural} calculate the shadow map without simultaneously optimizing shadow and other unknowns due to the discontinuity of the binary representation of the shadow map; the others either need the ground truth of shadow maps and ignore the shading cue~\cite{karnieli2022deepshadow}, or rely on priors of scene or light to ensure the accuracy~\cite{yang2022s,yang2022ps}.  
In contrast, DANI-Net simultaneously optimizes shadow and other unknowns, and exploits shadow and reflectance cues for UPS.

\textbf{Neural reflectance representation in 3D vision.}
Neural Radiance Fields (NeRFs)~\cite{mildenhall2020nerf} implicitly represent the shape information through MLPs. 
While NeRF~\mbox{\cite{mildenhall2020nerf}} applies volume rendering~\mbox{\cite{kajiya1984ray}} to optimize the MLPs through reconstruction loss, they often end up with `fake reflectance'~\mbox{\cite{verbin2021ref}} (\ie, the emitters inside the objects contribute to specific reflectance effects) and inaccurate geometries due to the lack of an implicit surface representation~\mbox{\cite{xie2022neural}}. To solve this, works like~\mbox{\cite{wang2021neus,oechsle2021unisurf}} combine surface rendering~\mbox{\cite{kajiya1986rendering}} with volume rendering~\mbox{\cite{kajiya1984ray}}. Other works~\mbox{\cite{verbin2021ref,zhang2021nerfactor,Zhang_2021_CVPR,Boss_2021_ICCV,Srinivasan_2021_CVPR}} adopt explicit representation on surface's reflectance to mimic realistic reflectance, which improves the accuracy of NeRF~\mbox{\cite{mildenhall2020nerf}}. 
Different from the above methods that work on multi-view inputs and focus on the full but coarse geometry recovery, DANI-Net takes single-view inputs under a UPS setup and focuses on the fine-detailed recovery of depth and surface normal. 


\begin{table*}[t]
\setlength{\abovecaptionskip}{0cm}
\setlength{\belowcaptionskip}{0cm}
\setlength{\tabcolsep}{15pt}
\caption{A summary of differences between the proposed DANI-Net and representative existing PS and UPS methods in terms of the solving problem, supervision, shadow handling strategy, and material model.  We categorize the shadow handling strategy as  `outliers' rather than `cues', if it models the shadow without explicitly exploiting it for shape recovery. `N.A.' represents Not Applicable. }
    \label{tab:method}
    \centering
    \vspace{0pt}
    \resizebox{1\linewidth}{!}{
    \begin{tabular}{c|cccc}
    \hline
    Method & Problem & Supervision & Shadow Handling & Material Model \\
    \hline
    TM18~\cite{taniai2018neural}  & PS    & Unsupervised  & Implicit (Outliers) & Isotropic (Bi-Polynomial BRDFs) \\
    LL22~\cite{li2022neural}  & PS    & Unsupervised  & Non-differentiable (Outliers) & Isotropic (MLP Bases) \\
        \hline
    CH19~\cite{chen2019self}  & UPS   &   Light + Surface Normal    & Implicit (Data-driven) & Implicit (Data-driven) \\
    CW20~\cite{chen2020learned}  & UPS   &    Light + Surface Normal   & Implicit (Data-driven) & Implicit (Data-driven) \\
    CM20~\cite{cho2018semi} & Semi-PS & Unsupervised & $\ell_1$-based Minimization (Outliers) & Lambertian \\
    SK22~\cite{sarno2022neural}  & UPS   &    Light + Surface Normal   & Implicit (Data-driven)  & Implicit (Data-driven) \\
    SCPS-NIR~\cite{li2022self} & UPS   & Unsupervised & Image Binarization (Outliers) & Isotropic (Gaussian Bases) \\
    \hline
    S3-NeRF~\cite{yang2022s} & Near Field PS &   Unsupervised    & Differentiable  (Cues for Invisible Shape) & Isotropic (Gaussian Bases) \\
    KF22~\cite{karnieli2022deepshadow}  & Near Field PS &   Unsupervised    & Known Shadow Maps  (Cues for Visible Shape) & N.A. \\ 
    \hline
    \textbf{DANI-Net}  & UPS   & Unsupervised & Differentiable  (Cues for Visible Shape) & Anisotropic (Gaussian Bases) \\
    \hline
    \end{tabular}
    }
    \vspace{-15pt}
\end{table*}

\section{Problem Definition}
\label{section_3}
Following a standard setup of UPS, we take input of a set of observed images $\boldsymbol{I} \triangleq (I_1, I_2, ..., I_f)$ of a static object captured under varying directional, parallel lights, and intend to recover light directions $\boldsymbol{L}\triangleq (\boldsymbol{l}_1, \boldsymbol{l}_2, ..., \boldsymbol{l}_f)$, light intensities $\boldsymbol{E}\triangleq (e_1, e_2, ..., e_f)$, and surface normal $\boldsymbol{N}\triangleq\{\boldsymbol{n}_i|i\in\mathbb{P}\}$ based on the observed images, where $\mathbb{P}$ is the set of all positions on the surface. The solution is achieved by solving the optimization problem, 

{\setlength\abovedisplayskip{0.cm}
\setlength\belowdisplayskip{0.1cm}
\begin{align}
\label{eq:opt}
\begin{aligned}
    \mathop{\arg\min}\limits_{\boldsymbol{L}, \boldsymbol{E}, \boldsymbol{N}}\sum_{i=1}^{\#\mathbb{P}}\sum_{j=1}^{f}{\text{D}(\bar{m}_{ij}, m_{ij})}, 
\end{aligned}
\end{align}}where 
$\bar{m}_{ij}\in I_j$ is the observed pixel intensity at position $i$, \#$\mathbb{P}$ is the number of elements in $\mathbb{P}$,
$\text{D}(\cdot,\cdot)$ is a metric to compute inverse rendering error, $m_{ij}$ is the corresponding rendered pixel intensity. 
With the linear radiometric response and an orthographic camera, $m_{ij}$ can be formulated as, 
{\setlength\abovedisplayskip{0.1cm}
\setlength\belowdisplayskip{0.1cm}
\begin{align}
\label{eq:general_brdf}
\begin{aligned}
    m_{ij} & =  e_j s_{ij} \rho_{ij}\max(\boldsymbol{n}_i^\top \boldsymbol{l}_j, 0)\\
           & = e_j s_{ij} (\rho^s_{ij}+ \rho^d_{ij})\max(\boldsymbol{n}_i^\top \boldsymbol{l}_j, 0),
\end{aligned}
\end{align}}where 
$\rho_{ij}=\rho(\boldsymbol{n}_i, \boldsymbol{l}_j, \boldsymbol{V}_d)$ describes the general reflectance which contains diffuse $\rho^d_i$ and specular $\rho^s_{ij}$ components, $\boldsymbol{V}_d=[0,0,1]$ is the view direction, $\max(\boldsymbol{n}_i^\top \boldsymbol{l}_j,0)$ describes the attached shadow, and $s_{ij}$ is the cast shadow calculated from the global shape and light. 

\textbf{Ambiguities in UPS.}
Unknown light in UPS brings the generalized bas-relief (GBR) ambiguity~\cite{belhumeur1999}, which can be represented by a $3\times 3$ matrix for the Lambertian surface.
For a non-Lambertian surface, specularity helps alleviate the GBR ambiguity~\cite{georghiades2003incorporating}. 
However, general reflectance introduces more unknowns, which require extra cues to solve. Another ambiguity is reflectance-light ambiguity denoted as a non-zero scalar between $e_j$ and other terms in \eref{eq:general_brdf}. 
Researchers focus on alleviating the GBR ambiguity while ignoring reflectance-light ambiguity, as the former determines the accuracy of the estimated surface normal. This paper also focuses on alleviating GBR ambiguity.

\section{Proposed DANI-Net}
This section elaborates on how DANI-Net solves UPS. 
We first present a differentiable shadow handling method that fully exploits shadow cues to solve UPS and introduces an anisotropic reflectance model that describes a more precise and realistic specularity to improve performance on general materials. 
We then show how the proposed DANI-Net connects the differentiable shadow handling method, the anisotropic reflectance model, and the inverse rendering together to jointly optimize the shape and light in an unsupervised manner.
Finally, we introduce additional loss functions to train DANI-Net.

\begin{figure}[t]
    \centering
    \includegraphics[width=0.46\textwidth]{"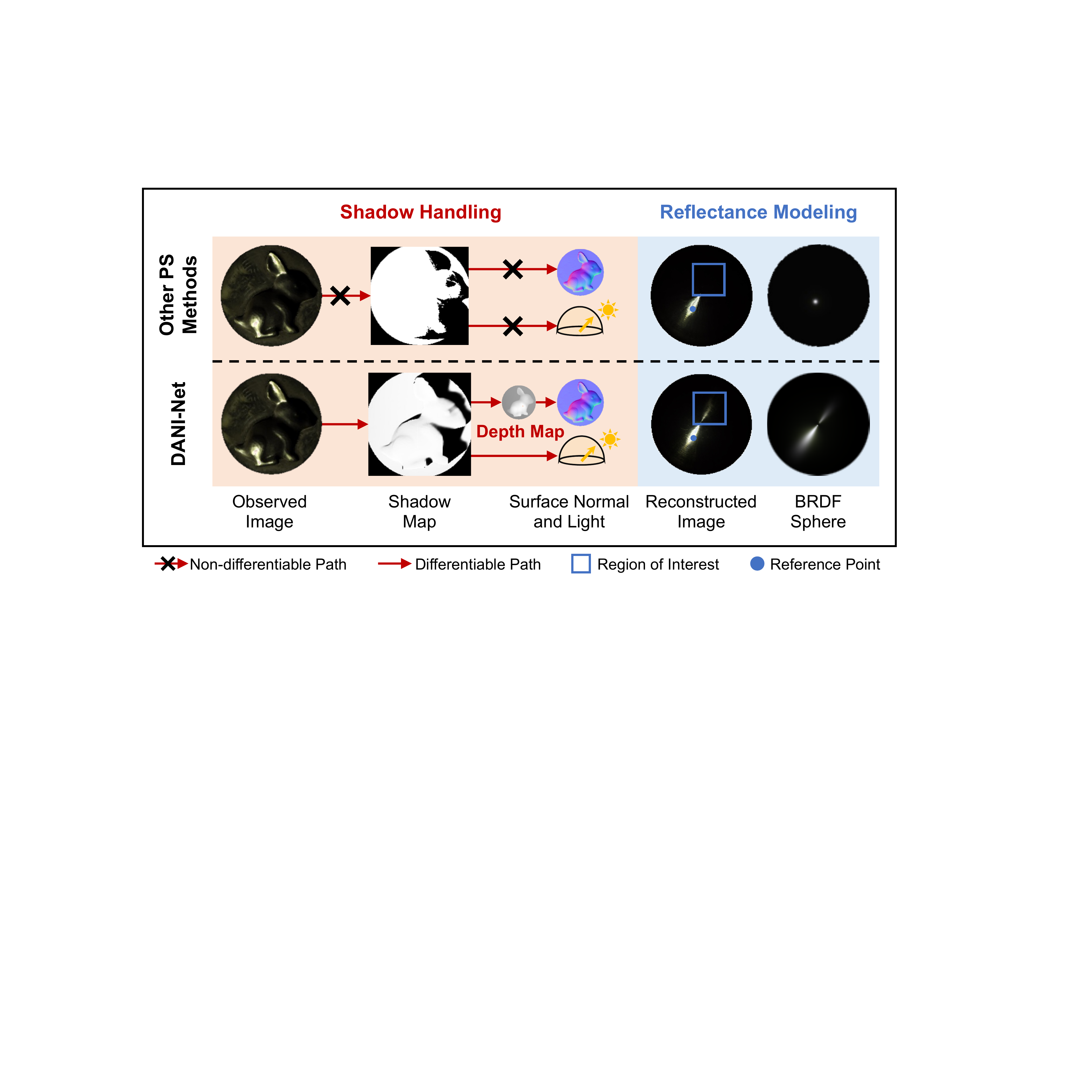"} 
    \vspace{-7pt}
    \caption{An illustration of our shadow calculation methods, which shows how sampled points and surface points are computed, based on which the shadow is calculated by \Eref{eq:shadow}. } 
    \label{fig:interpolate}
    \vspace{-14pt}
\end{figure}

\subsection{Differentiable Shadow Handling}
\label{sec:shadow}

Finding the differentiable relationship between shadow, shape, and light is beneficial for solving UPS since 1) it generates a more accurate shadow map through iterative optimization which leads to a more accurate shape-light estimation; 2) compared to previous PS/UPS methods~\cite{li2022neural,li2022self} lacking usage of global shape-light information, it facilitates fully exploiting shadow cues containing that information to solve UPS.
While the attached shadow is already differentiable to the surface normal and light direction in a specific domain, finding the differentiable relationship for cast shadow is more tricky, since it directly associates with light and the entire depth map instead of the surface normal at a particular point. Such an association raises two key technical challenges to optimizing DANI-Net. First, since the pixel-wise cast shadow calculation increases the computational cost, it is necessary to introduce a technique to improve the training efficiency. Second, since there is no direct connection between the cast shadow and the surface normal, an effective fitting method that calculates the surface normal from the depth map of limited resolution is another required technique to ensure the backpropagation efficiency and the connection from the depth map to the reflectance model for inverse rendering.

\textbf{Shadow calculation.}
The basic idea of our shadow calculation method is to consider the occlusion of a reference point according to a few points rather than all points.
As shown in \Fref{fig:interpolate}, given the depth map  $W\triangleq \{ w_{i} | 1 \leq i \leq \#\mathbb{P}\}$, the `soft shadow'~\cite{liu2019soft, pagurekdifferentiable} $s_{ij}$ of a reference point $\boldsymbol{p}_i$ illuminated by a light with direction $\boldsymbol{l}_j$ is calculated as:
{\setlength\abovedisplayskip{0.2cm}
\setlength\belowdisplayskip{0.2cm}
\begin{align}{\small
\begin{aligned}
    \label{eq:shadow}
    s_{ij} = \text{Sigmoid}(\alpha(\min\{w^k_{ij}-\hat{w}^k_{ij}|1 \leq k \leq N_p\}) + \beta),
    \end{aligned}}
\end{align}
}where 
$\{\hat{w}^k_{ij}|1 \leq k \leq N_p\}$ are depth values of {\it sampled points} $\{\boldsymbol{p}_{ij}^{k}|1 \leq k \leq N_p\}$ uniformly sampled from the {\it light segment} that starts at $\boldsymbol{p}_i$ along $\boldsymbol{l}_j$ and ends at $\boldsymbol{p}_{ij}^{N_p}$\footnote{$\boldsymbol{p}_{ij}^{N_p}$'s projection in xy-plane lies on the boundary, shown in \Fref{fig:interpolate}.}. 
$\{{w}^k_{ij}|1 \leq k \leq N_p\}$ are depth values on the surface with the same xy-coordinates to sampled points.
$N_p$ is the number of sampled points set as 64 in our implementation.
The Sigmoid function and learnable parameters of $\alpha$ and $\beta$ are adopted to calculate the `soft shadow'. 
In practice, ${w}^k_{ij}$ cannot be directly retrieved from $W$ given the limited resolution of $W$, as shown in \Fref{fig:interpolate}.
Although we can directly infer $w^k_{ij}$ through a Multi-layer Perceptron (MLP), it significantly increases the training time.
To this end, we apply {\it grid bi-linear interpolation} to interpolate $w^k_{ij}$ by retrieving its four neighboring points' depth values from $W$, as shown in \Fref{fig:interpolate}.  The grid bi-linear interpolation is much faster\footnote{The inference time of grid bi-linear interpolation vs. MLP inference is 0.002 vs. 0.212 seconds on average, for 64 sampled points and a batch size of 1 on {\sc DiLiGenT} dataset~\cite{shi2016benchmark}} than inferring $w^k_{ij}$, which speeds up the training process.

\begin{figure}[t]
    \centering
    \includegraphics[width=0.46\textwidth]{"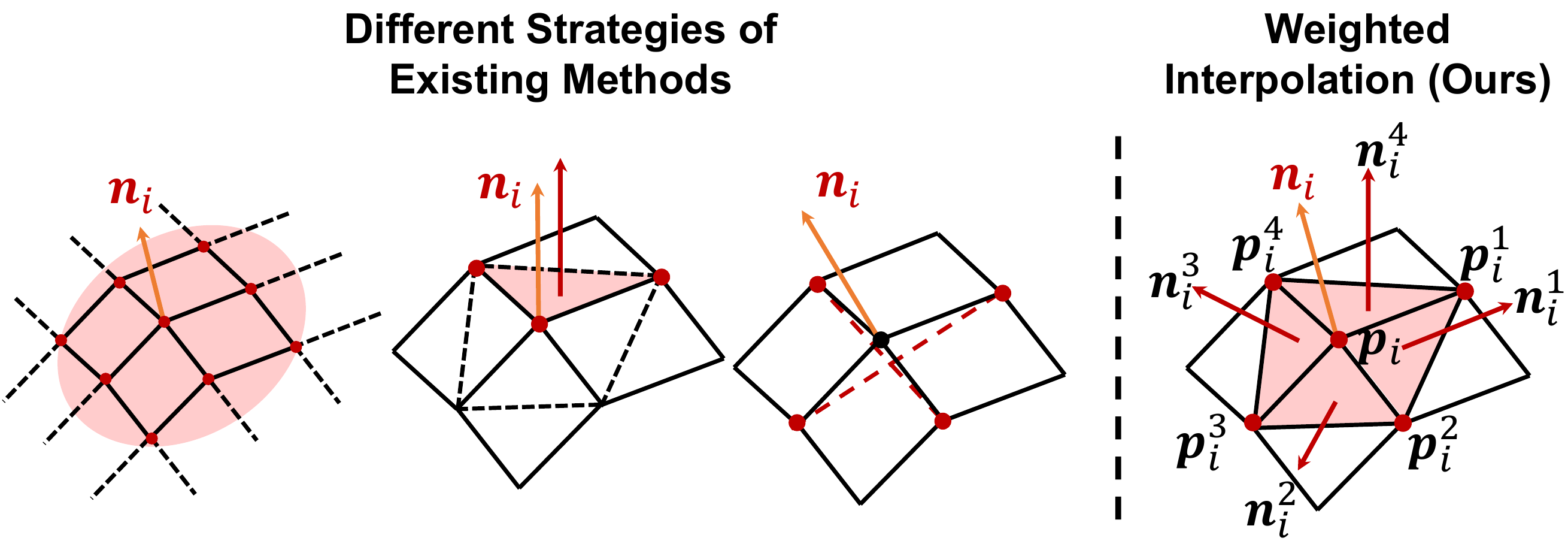"} 
    \vspace{-10pt}
    \caption{An illustration of different strategies of existing methods and the proposed surface normal fitting method (\Eref{eq:norm_cal}). We mark the points, triangles, or lines that are leveraged to fit the surface normal in red for easy comparison. From left to right, method that uses excessive neighbor points~\cite{qi2018geonet}, just a few points~\cite{nakagawa2015estimating}, a proper number of neighbor points while ignoring the information of query point~\cite{li2022neural}, and our method that jointly considers a proper number of neighbor points and the query point.}
    \label{fig:fit}
    \vspace{-15pt}
\end{figure}

\textbf{Surface normal fitting.} 
The surface normal fitting method in this work requires pixel-level accuracy and deals with the depth map that is dynamically optimized according to the backpropagation from fitting results.
Existing solutions cannot be directly applied because 
they either use excessive neighbor points ($> 4$)~\cite{qi2018geonet, nehab2005efficiently} that reduces the precision of the fitted surface normal, or use just a few points~\cite{nakagawa2015estimating} ($<4$) that brings ineffective backpropagation since the gradients affect a limit number of points on the depth map, or use a proper number of points ($=4$) while ignoring the information of the query point~\cite{li2022neural}, as shown in \Fref{fig:fit}.
To solve these problems, we propose a weighted interpolation method to fit the surface normal that jointly considers the depth information of the query point and a proper number of neighbor points.
As shown in \Fref{fig:fit}, our method first computes normal vectors $\{\boldsymbol{n}_i^k|k=1,2,3,4\}$ of query's adjacent triangles and then fit the normal $\boldsymbol{n}_i$ of the query point $\boldsymbol{p}_i$ by weighted interpolation of these vectors,
{\setlength\abovedisplayskip{0.2cm}
\setlength\belowdisplayskip{0.2cm}
\begin{align}{\footnotesize
\begin{aligned}
\label{eq:norm_cal}
    \boldsymbol{n}_i & = \sum_{k=1}^{4}\gamma^k_i \boldsymbol{n}^k_i =\sum_{k=1}^{4}\gamma^k_i\operatorname{Nor}[(\boldsymbol{p}^{k+1}_i - \boldsymbol{p}_i) \times (\boldsymbol{p}_i^{k} - \boldsymbol{p}_i)]^{\top}, \\
    \gamma^k_i & = \frac{|d^k_i|^{-1}}{\sum_{k=1}^{4}|d^k_i|^{-1}},~~~d^k_i = w^k_i + w_i^{k+1} - 2 w_i,\\
    \end{aligned}}
\end{align}}where 
$\boldsymbol{p}_i^k$ are adjacent points of $\boldsymbol{p}_i$, 
$k=1$ if $k+1 > 4$, `$\operatorname{Nor}$' is the vector normalization operation, $w^k_i$ is the depth value of point $\boldsymbol{p}_i^k$. In this way, our method generates a high-precision normal map and backpropagates gradients to a proper number of points for the depth map optimization\footnote{A further comparison between our weighted interpolation method and other normal fitting methods can be found in the supplementary material.}.

\begin{figure}[t]
    \centering
    \includegraphics[width=0.46\textwidth]{"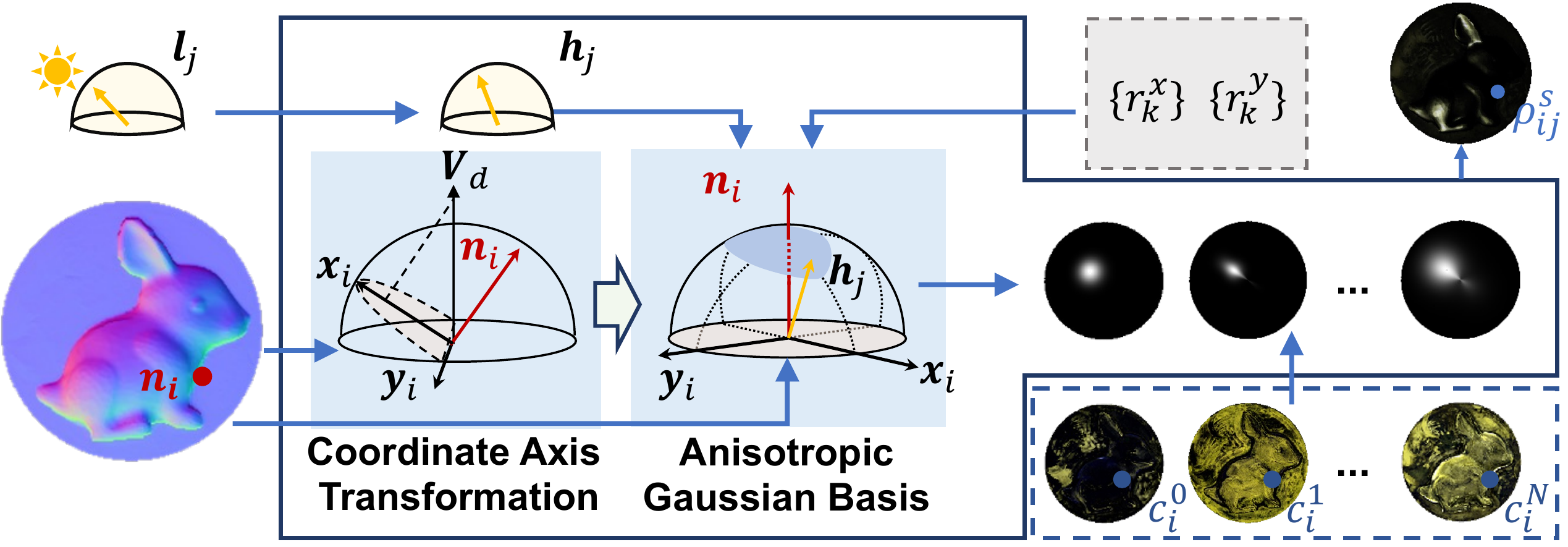"} 
    \vspace{-5pt}
    \caption{An illustration of building up the anisotropic reflectance model by \Eref{eq:asg}. A group of ASG bases with various specular lobe's widths ($r^x_k$ and $r^y_k$)~\cite{walter2005notes} in the direction of tangent vector $\boldsymbol{x}_i$ and binormal vector $\boldsymbol{y}_i$ are multiplied by the corresponding spatially varying weight $c^k_i$ to compute $\rho^s_{ij}$.} 
    \label{fig:material}
    \vspace{-15pt}
\end{figure}
\subsection{Anisotropic Reflectance Modeling}
We build up a more sophisticated and precise reflectance model to optimize the shape and light through anisotropic reflectance. As illustrated in \Fref{fig:material}, we represent the spatially varying anisotropic specularity as a weighted sum of Anisotropic Spherical Gaussian (ASG)~\cite{xu2013anisotropic} bases\footnote{We neglect the Fresnel term applied in micro-facet reflectance models (\eg, Ward model~\cite{goldman2009shape}) for easier convergence.}.
{\setlength\abovedisplayskip{0.1cm}
\setlength\belowdisplayskip{0.1cm}
\begin{align}
\begin{aligned}
\label{eq:asg}
    \rho_{ij}^s &= \sum_{k=1}^{N_G} c_i^{k} [e^{-r_k^x(\boldsymbol{h}_j\cdot \boldsymbol{x}_i)^2 - r_k^y(\boldsymbol{h}_j\cdot \boldsymbol{y}_i)^2}],\\
\end{aligned}
\end{align}}where 
$k \in [1, 2, ..., N_G]$. $N_G$ is the number of the ASG bases empirically set as 12. $c_i^k$ is spatially varying weights that balance different ASG bases to model spatially varying specularity. 
$\boldsymbol{h}_j=\frac{\boldsymbol{V}_d+\boldsymbol{l}_j}{\|\boldsymbol{V}_d+\boldsymbol{l}_j\|}$ is the half-unit-vector between view direction $\boldsymbol{V}_d$ and light direction $\boldsymbol{l}_j$.
$r_k^x$ and $r_k^y$ are the lobe's width in the direction of $\boldsymbol{x}_i$ and $\boldsymbol{y}_i$, respectively.  When $r_k^x = r_k^y$, ASG degrades to Isotropic Spherical Gaussian (ISG) bases. 
$\boldsymbol{x}_i$ and $\boldsymbol{y}_i$ are the tangent and binormal vectors, respectively, of the surface tangent plane at point $\boldsymbol{p}_i$. 
As shown in \Fref{fig:material}, $\boldsymbol{x}_i$ and $\boldsymbol{y}_i$ are calculated by:
{\setlength\abovedisplayskip{0.1cm}
\setlength\belowdisplayskip{0.1cm}
\begin{align}
\begin{aligned}
\label{eq:xy}
    \boldsymbol{x}_i = \boldsymbol{V}_d - (\boldsymbol{V}_d \cdot \boldsymbol{n}_i) \boldsymbol{n}_i,~~~
    \boldsymbol{y}_i = \boldsymbol{n}_i \times \boldsymbol{x}_i.
\end{aligned}
\end{align}
The annealing strategy in~\cite{li2022self} is adopted to control the number of activated ASG bases at different training stages, which helps avoid local optimum at the beginning. 
}
\subsection{Neural Inverse Rendering}
\begin{figure}[t]
    \centering
    \includegraphics[width=0.46\textwidth]{"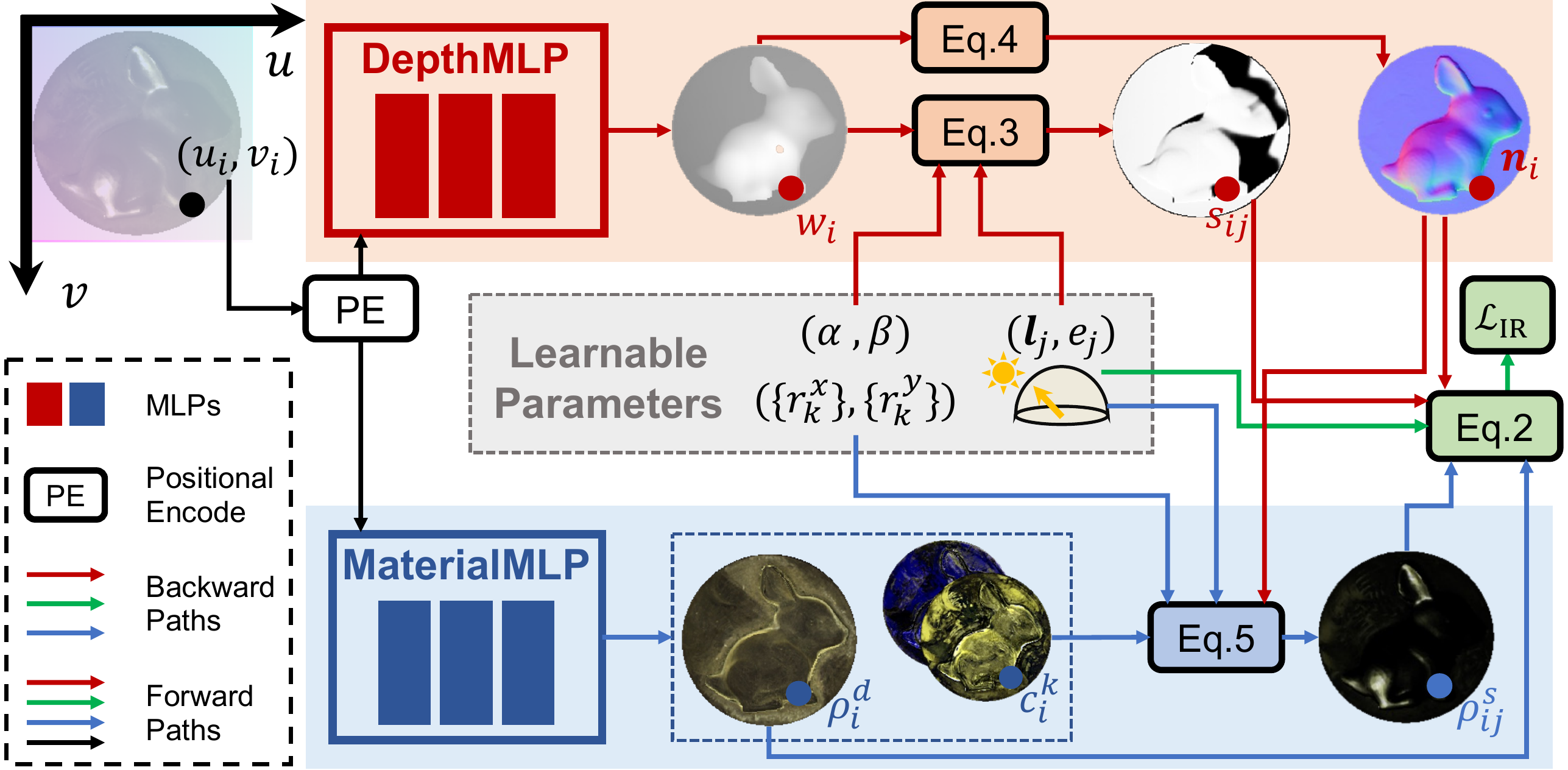"} 
    \vspace{-7pt}
    \caption{Framework overview of DANI-Net. DepthMLP takes the positional code as the input and outputs depth $w_i$. MaterialMLP takes the same positional code as the input and outputs $\rho_i^d$ and $c_i^k$. The shadow $s_{ij}$ is calculated based on \Eref{eq:shadow}. Spatially varying anisotropic specularity $\rho^s_{ij}$ is obtained through \Eref{eq:asg}. 
    Rendered images are generated through \Eref{eq:general_brdf}. 
    The inverse rendering loss $\mathcal{L}_\text{IR}$ measures the rendering error between rendered images and observed images, and backpropagates to the light calibration and shape estimation through the differentiable shadow path (in red and green colors) and the anisotropic reflectance path (in blue and green colors). } 
    \label{fig:overview}
    \vspace{-12pt}
\end{figure}

As shown in \Fref{fig:overview}, our neural inverse rendering framework integrates the differentiable shadow handling method and anisotropic reflectance model to two MLPs\footnote{More implementation details, such as the MLPs' structures, could be found in the supplementary material.}
 and several learnable parameters. 

\textbf{DepthMLP and MaterialMLP.} 
To consider spatially-dependent information, we build up a neural depth field named {\it DepthMLP} and a neural material field named {\it MaterialMLP}, implemented by two MLPs.
The DepthMLP estimates $w_{i}$ pixel-wisely; the MaterialMLP outputs $c^k_i$ and $\rho^d_i$ for spatially varying reflectance.
We apply the same method~\cite{mildenhall2020nerf} to encode the coordinate on the image plane.

\textbf{Learnable parameters.} 
As shown in \Fref{fig:overview}, light conditions (\ie, $\boldsymbol{l}_j$, $e_j$), ASG lobe’s width parameters (\ie, $r^x_k$, $r^y_k$), and soft shadow parameters (\ie $\alpha$ and $\beta$) are set as learnable parameters. 
We initialize $\boldsymbol{l}_j$ and $e_j$ using outputs of a pre-trained light calibration model, \ie, we apply the CNN structures in~\cite{li2022self} and train on the Blobby and Sculpture dataset~\cite{chen2018ps}.
We set the scope of $r^x_k$ and $r^y_k$ in $[10^0, 10^3]$ and initialize $r^x_k=r^y_k=10^{\frac{(\log300-\log10)(k-1)}{N_G-1}+\log10}$ (\ie, uniformly sampled ${N_G}$ values from $[10,300]$ in the log space) to force an isotropic reflectance model and consider different sharpness at the beginning, which enables a smooth transition to the anisotropic reflectance. 
$\alpha$ and $\beta$ are empirically initialized to 400 and 3.

A rendered image is calculated through \Eref{eq:general_brdf} using the outputs from the two MLPs and the learnable parameters. 
The inverse rendering $\ell_1$ metric error $\mathcal{L}_{\text{IR}}$ between the rendered images and the observed images given by \Eref{eq:reconstruct} passes through the differentiable paths to jointly update weights of MLPs and learnable parameters,
{\setlength\abovedisplayskip{0.1cm}
\setlength\belowdisplayskip{0.1cm}
\begin{equation}
\label{eq:reconstruct}\footnotesize
        \mathcal{L}_{\text{IR}} = \frac{1}{\text{\#} \mathbb{P} \times f}\sum_{i=1}^{\text{\#}\mathbb{P}}\sum_{j=1}^{f}{|\bar{m}_{ij} - e_js_{ij}(\rho^d_i+\rho^s_{ij})\max(\boldsymbol{n}_i^\top \boldsymbol{l}_j, 0)|}.
\end{equation}
}

\begin{table*}[t]
    \setlength{\tabcolsep}{10pt}
    \caption{Quantitative comparison in terms of MAE of surface normal  on {\sc DiLiGenT} benchmark dataset~\cite{shi2016benchmark}. \textbf{Bold numbers} and \underline{underlined numbers} indicate the best and the second-best results among UPS methods, respectively.}
    \vspace{-10pt}
    \label{tab:normal_diligent}
    \centering
    \resizebox{1\linewidth}{!}{
    \begin{tabular}{c|cccccccccc|c}
    \hline
    Method & \sc{Ball}  & \sc{Bear}  & \sc{Buddha} & \sc{Cat}   & \sc{Cow}   & \sc{Goblet} & \sc{Harvest} & \sc{Pot1}  & \sc{Pot2}  & \sc{Reading} & AVG    \\ \hline
    LS~\cite{woodham1980} & 4.10  & 8.39  & 14.92 & 8.41  & 25.60  & 18.50   & 30.62 & 8.89  & 14.65 & 19.80  & 15.39  \\
    TM18~\cite{taniai2018neural} & 1.47  & 5.79  & 10.36 & 5.44  & 6.32  & 11.47  & 22.59 & 6.09  & 7.76  & 11.03 & 8.83   \\
    LL22~\cite{li2022neural} & 2.43 & 3.64 & 8.04 & 4.86 & 4.72 & 6.68 & 14.90 & 5.99 & 4.97 & 8.75 & 6.50 \\
    \hline
    \hline
    PF14\cite{papadhimitri2014closed} & 4.77  & 9.07  & 14.92 & 9.54  & 19.53 & 29.93  & 29.21 & 9.51  & 15.90 & 24.18 & 16.66 \\
    CH19~\cite{chen2019self} & 2.77  & 6.89  & 8.97  & 8.06  & 8.48  & 11.91  & 17.43 & 8.14  & 7.50  & 14.90 & 9.51  \\
    CW20~\cite{chen2020learned} & 2.50  & 5.60  & \textbf{8.60}  & 7.90  & 7.80  & 9.60   & 16.20 & 7.20  & 7.10  & 14.90 & 8.71  \\
    SCPS-NIR~\cite{li2022self} & \textbf{1.24} & \textbf{3.82} & 9.28 & \textbf{4.72} & 5.53 & 7.12 & \underline{14.96} & \underline{6.73} & 6.50 & 10.54 & 7.05 \\
    \hline
    DANI-Net {\it w/o} $s$ & 1.71  & \underline{3.95}  & 8.71  & 4.95  & \textbf{4.95}  & \textbf{6.80}   & 16.00    & 7.04  & \textbf{5.27}  & \underline{9.32}  & \underline{6.87} \\
    DANI-Net {\it w} \cite{li2022neural} &  \underline{1.64}  & 4.03  & 9.16  & 5.27  & \underline{5.22}  & 6.98  & 16.43  & 6.85  & 5.52  & 9.53  & 7.06 \\
    DANI-Net & 1.65   & 4.11  & \underline{8.69}  & \underline{4.73}  & 5.52  & \underline{6.96}  & \textbf{13.99} & \textbf{6.41}  & \underline{5.29}  & \textbf{8.08}  & \textbf{6.54} \\
    \hline
    \end{tabular}
    }
    \vspace{-15pt}
\end{table*}
\subsection{Optimizing DANI-Net}

In addition to the inverse rendering loss function, the proposed DANI-Net is optimized by the silhouette loss function $\mathcal{L}_{\text{Si}}$ and the smooth loss function.
The silhouette loss function $\mathcal{L}_\text{Si}$ is similar to those in~\cite{hashimoto2019uncalibrated, chen2020learned, li2022self} by calculating the cosine similarity between estimated and fitted silhouette's normal\footnote{Fitted silhouette's normal is pre-computed based on the geometry intuition, \ie, the projection of silhouette's normal to xy-plane is likely to be perpendicular to the silhouette if we assume the silhouette is occluding. For objects with non-occluding silhouette, we apply a flexible strategy to use $\mathcal{L}_{\text{Si}}$, please refer to the supplementary material for more details.}.
The fitted surface normals at the silhouette alleviate the GBR ambiguity because the GBR transformation on these surface normals equals an identity matrix. 
Our smoothness loss function focuses on the diffuse reflectance map $R^d$, the normal map $\boldsymbol{N}$, and the depth map $W$, which follows the similar implementation in~\cite{li2022self, li2022neural},
{\setlength\abovedisplayskip{0.1cm}
\setlength\belowdisplayskip{0.1cm}
\begin{align}{\footnotesize
\begin{aligned}
     \mathcal{L}_{\text{smooth}} =& \lambda \mathcal{L}_{R^d}+ \lambda \mathcal{L}_{W}+\lambda_N \mathcal{L}_N\\
    = & \lambda \frac{1}{\#\mathbb{P}} \sum_{i=1}^{\#\mathbb{P}}|\frac{\partial R^d}{\partial u} + \frac{\partial R^d}{\partial v}|
     +  \lambda \frac{1}{\#\mathbb{P}} \sum_{i=1}^{\#\mathbb{P}}|\frac{\partial W}{\partial u} + \frac{\partial W}{\partial v}|\\
   + &\lambda_N \frac{1}{\#\mathbb{P}} \sum_{i=1}^{\#\mathbb{P}}|\frac{\partial \boldsymbol{N}}{\partial u} + \frac{\partial \boldsymbol{N}}{\partial v}|.
    \end{aligned}}
\end{align}
}

We train the DANI-Net in three stages\footnote{Please refer to the supplementary material for more ablation studies about our three-stage training schema and silhouette loss.} for fast convergence.
The loss function in three stages is as follows,
{\setlength\abovedisplayskip{0.1cm}
\setlength\belowdisplayskip{0.1cm}
\begin{align}
\begin{aligned}
      \mathcal{L}_\text{stage1}=&\mathcal{L}_\text{IR}+ \lambda_\text{Si} \mathcal{L}_\text{Si}+ \mathcal{L}_{\text{smooth}},\\
       \mathcal{L}_\text{stage2}=&\mathcal{L}_\text{IR}  + \lambda_\text{Si}\mathcal{L}_\text{Si}+ \lambda\mathcal{L}_N,\\
       \mathcal{L}_\text{stage3}=&\mathcal{L}_\text{IR} + \lambda_\text{Si}\mathcal{L}_\text{Si},
    \end{aligned}
\end{align}
}where 
$\lambda=0.01$, $\lambda_N=0.02$, and $\lambda_\text{Si}=0.01$.
Three stages take 500, 1000, and 500 epochs, respectively.
We also apply a progressive training schema with the learning rate decaying in a cosine annealing manner. 
During training, we use Adam as the optimizer with a learning rate $\alpha_l=0.001$ decaying to $0.0001$ for 2000 epochs.

\section{Experiments}
\textbf{Evaluation Metrics.} To evaluate the accuracy of the predicted light directions and surface normal, we adopt mean angle error (MAE) in degree. 
To evaluate the accuracy of recovered light intensity free from the reflectance-light ambiguity, we use the scale-invariant relative error~\cite{chen2019self}, $E_\text{int}=\frac{1}{f} \sum_{j=1}^{f}\left(\frac{\left|\eta e_{j}-\tilde{e}_{j}\right|}{\tilde{e}_{j}}\right)$, where $\tilde{e}_{j}$ is the ground truth light intensity, $e_{j}$ is the estimated light intensity, and $\eta$ is calculated through solving $\mathop{\arg\min} _\eta\sum_{j=1}^{f}\left(\eta e_{j}-\tilde{e}_{j}\right)^{2}$ by least squares.

\textbf{Datasets.} We evaluate our method on {\sc DiLiGenT}~\cite{shi2016benchmark} and {\sc DiLiGenT}$10^2$ benchmark dataset~\cite{ren2022diligent102}\footnote{Results on {\sc Light Stage Data Gallery} dataset~\cite{chabert2006relighting} and {\sc Apple \& Gourd} dataset~\cite{alldrin2008photometric} can be found in the supplementary material.}.

\subsection{Performance on {\sc DiLiGenT}~\cite{shi2016benchmark}}

We compare with six state-of-the-art photometric stereo methods on {\sc DiLiGenT} dataset~\cite{shi2016benchmark}, including two unsupervised calibrated PS methods (TM18~\cite{taniai2018neural} and LL22~\cite{li2022neural}), two supervised UPS methods (CH19~\cite{chen2019self} and CW20~\cite{chen2020learned}), and an unsupervised UPS method (SCPS-NIR~\cite{li2022self}). We also include two baseline methods for PS (LS~\cite{woodham1980}) and UPS (PF14~\cite{papadhimitri2014closed}).

\textbf{Surface normal estimation.}
\Tref{tab:normal_diligent} shows that DANI-Net outperforms other UPS methods and achieves comparable performance as the state-of-the-art calibrated PS method LL22~\cite{li2022neural}\footnote{Visual quality comparison on {\sc DiLiGenT} dataset~\cite{shi2016benchmark} can be found in the supplementary material.}.
For {\sc Reading} and {\sc Harvest} that contain substantial cast shadow, DANI-Net achieves a superior performance advantage over the other methods because our differentiable handling method simultaneously optimizes the shadow map with other unknowns, ensuring full utilization of shadow cues. For {\sc Cow} and {\sc Goblet} with a more prominent anisotropic reflectance as compared to other objects, DANI-Net also achieves the best performance, validating the effectiveness of our anisotropic material model.

\begin{table*}[t]
    \centering
    \setlength{\abovecaptionskip}{0cm}
    \setlength{\belowcaptionskip}{0cm}
    \caption{Quantitative comparison in terms of MAE of light direction and scale-invariant error of intensity on {\sc DiLiGenT} benchmark dataset~\cite{shi2016benchmark}. \textbf{Bold numbers} and \underline{underlined numbers} indicate the best and the second-best results, respectively. }
    \label{tab:light_diligent}
    \resizebox{\linewidth}{!}{
    \begin{tabular}{c|cccccccccccccccccccc|cc}
    \hline
    \multirow{2}[2]{*}{Model}& \multicolumn{2}{c}{\sc{Ball}} & \multicolumn{2}{c}{\sc{Bear}} & \multicolumn{2}{c}{\sc{Buddha}} & \multicolumn{2}{c}{\sc{Cat}} & \multicolumn{2}{c}{\sc{Cow}} & \multicolumn{2}{c}{\sc{Goblet}} & \multicolumn{2}{c}{\sc{Harvest}} & \multicolumn{2}{c}{\sc{Pot1}} & \multicolumn{2}{c}{\sc{Pot2}} & \multicolumn{2}{c|}{\sc{Reading}} & \multicolumn{2}{c}{AVG} \\ 
    & dir. & int.  & dir. & int.  & dir. & int.  & dir. & int.  & dir.  & int.  & dir.  & int.  & dir.  & int.  & dir. & int.  & dir.  & int.  & dir.  & int.  & dir.  & int.  \\
    \hline
    PF14~\cite{papadhimitri2014closed}          & 4.90 & 0.036 & 5.24 & 0.098 & 9.76 & 0.053 & 5.31 & 0.059 & 16.34 & 0.074 & 33.22 & 0.223 & 24.99 & 0.156 & \underline{2.43} & \textbf{0.017} & 13.52 & \textbf{0.044} & 21.77 & 0.122 & 13.75 & 0.088 \\
    CH19~\cite{chen2019self}         & 3.27 & 0.039 & 3.47 & 0.061 & 4.34 & 0.048 & 4.08 & 0.095 & 4.52  & 0.073 & 10.36 & 0.067 & 6.32  & 0.082 & 5.44 & 0.058 & 2.87  & \underline{0.048} & \underline{4.50}   & 0.105 & 4.92  & 0.068 \\
    CW20~\cite{chen2020learned}         & 1.75 & 0.027 & 2.44 & 0.101 & 2.86 & 0.032 & 4.58 & 0.075 & 3.15  & \textbf{0.031} & 2.98  & \textbf{0.042} & \underline{5.74}  & 0.065 & \textbf{1.41} & 0.039 & \underline{2.81}  & 0.059 & 5.47  & 0.048 & \underline{3.32} & 0.052 \\
    SCPS-NIR~\cite{li2022self} & 1.43 & \textbf{0.019} & \textbf{1.56} & \textbf{0.019} & 4.22 & \textbf{0.021} & 4.41 & 0.032 & 4.94 & 0.062 & \underline{2.26} & \textbf{0.042} & 6.41 & \textbf{0.023} & 3.46 & 0.030 & 4.19 & 0.082 & 7.34 & 0.035 & 4.02 & \underline{0.037} \\
    \hline
    DANI-Net {\it w/o $s$} & 1.25  & \underline{0.020}  & 2.08  & 0.026 & \textbf{2.50}   & \textbf{0.021} & \textbf{3.09}  & \textbf{0.027} & \textbf{2.12}  & 0.059 & 2.89  & 0.047 & \textbf{5.27}  & \underline{0.026} & 4.23  & 0.031 & \textbf{2.76}  & 0.086 & 5.45  & \textbf{0.026} & \textbf{3.16}  & \underline{0.037} \\
     DANI-Net {\it w}\cite{li2022neural} & \underline{1.24}  & \underline{0.020}  & \underline{1.69}  & 0.024 & 3.07   & \textbf{0.021} & 3.28  & \underline{0.026} & \underline{2.41}  & 0.058 & 3.01  & \underline{0.043} & 6.47  & 0.030 & 4.50  & \underline{0.029} & 3.30 & 0.084 & 5.37  & \textbf{0.025} & 3.43  & \textbf{0.036} \\
    DANI-Net & \textbf{1.23}  & \underline{0.020}  & 3.71  & \underline{0.022}  & \underline{2.63}  & \underline{0.025}  & \underline{3.32}  & 0.029  & 4.19  & \underline{0.055}  & \textbf{1.65}  & 0.044  & 6.34  & \underline{0.026}  & 4.17  & \underline{0.029}  & 3.42  & 0.079  & \textbf{3.28}  & \underline{0.028}  & 3.39  & \textbf{0.036}  \\
    \hline
    \end{tabular}
    }
\vspace{-10pt}
\end{table*}

\begin{figure*}[t]
    \centering
    \includegraphics[width=0.98\textwidth]{"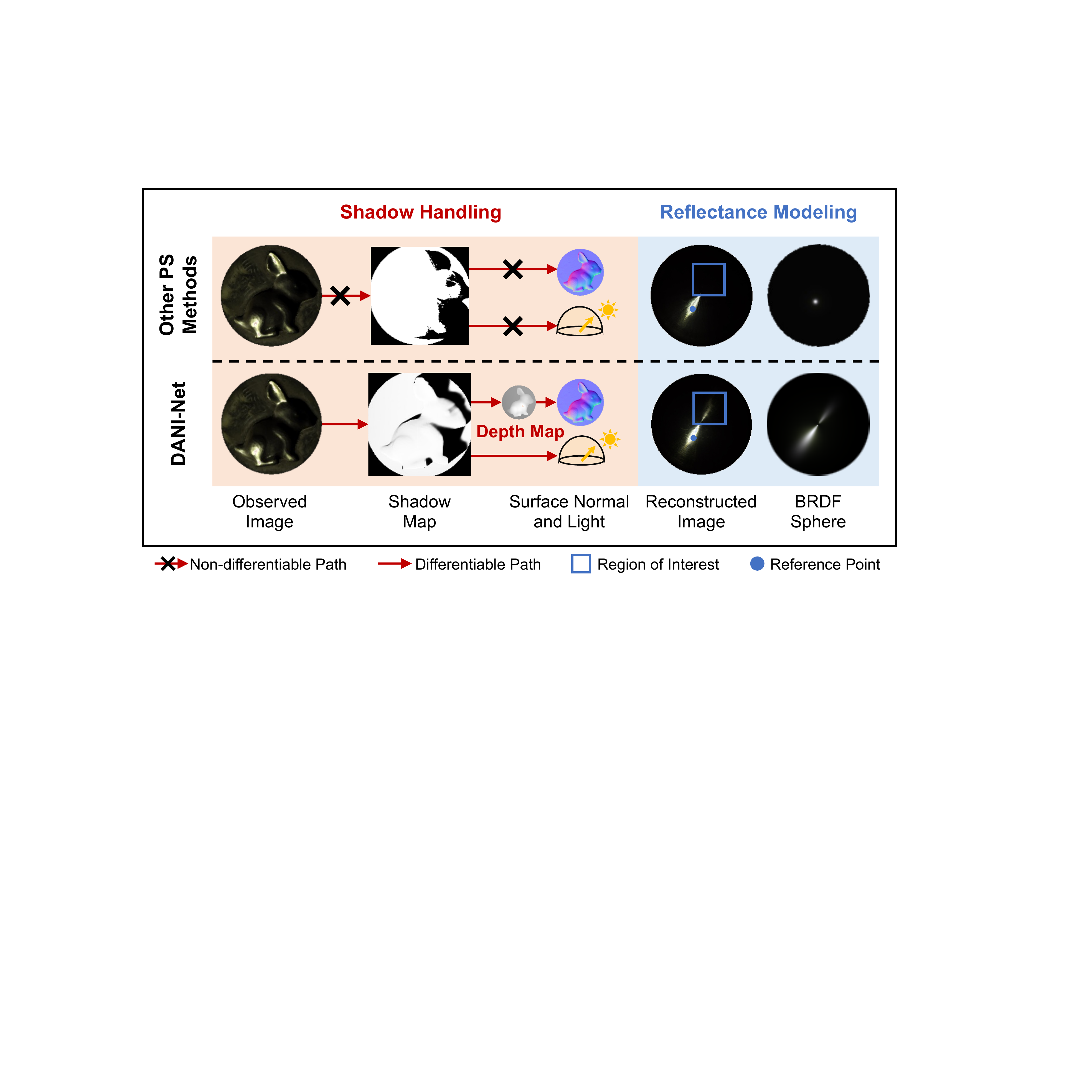"}
    \vspace{-10pt}
    \caption{A visual quality comparison on shadow maps (cast and attached shadow) and surface normal among LL22~\cite{li2022neural}, SCPS-NIR~\cite{li2022self}, DANI-Net {\it w/o s},  DANI-Net {\it w} \cite{li2022neural}, and DANI-Net on {\sc Harvest} from {\sc DiLiGenT}~\cite{shi2016benchmark}. 
    The four left columns are images/shadow maps under four light directions, labeled by ABCD, indicated by red dots shown in the light direction figure on the top right.
    The two right columns are the ground truth or estimated normal maps and error maps. Red boxes highlight prominent regions for easy comparison. } 
    \label{fig:shadow_comp}
    \vspace{-13pt}
\end{figure*}

\textbf{Light calibration.} 
As shown in \Tref{tab:light_diligent}, our method achieves the best light intensity and the second-best light direction calibration results. 
Interestingly, although CW20~\cite{chen2020learned} achieves better light direction calibration results than DANI-Net, it achieves a less comparable performance of surface normal estimation.
The reason could be that CW20~\cite{chen2020learned} and other two-stage methods suffer from the accumulating error brought by the light calibration stage.

\textbf{Differentiable shadow handling validation.} 
To validate the effectiveness of our differentiable shadow handling method, we compare it with `DANI-Net {\it w/o $s$}' and `DANI-Net {\it w} \cite{li2022neural}', which are two alternatives of DANI-Net without differentiable shadow handling.
As compared with DANI-Net, the only difference of the two alternatives are shadow handling method (\ie, DANI-Net {\it w/o $s$} uses the same fixed shadow map in SCPS-NIR~\cite{li2022self}; DANI-Net {\it w}\cite{li2022neural} uses same shadow handling method in~\cite{li2022neural}).
As can be observed in \Tref{tab:normal_diligent} and \Tref{tab:light_diligent}, the performance of DANI-Net {\it w/o $s$} and DANI-Net {\it w} \cite{li2022neural} drops on surface normal estimation (the average MAE increases $0.33^\circ$ and $0.52^\circ$, respectively), while remains similar on light calibration compared with DANI-Net. However, as shown in \Fref{fig:shadow_comp}, DANI-Net generates much more realistic and smooth shadow maps than its alternatives. 
This is because the shadow maps' calculation in \cite{li2022self,li2022neural} suffers from certain limitations. For~\cite{li2022self}, shadow maps are calculated based on pixel intensities rather than depth and light which can be affected by surface points with dark color; for~\cite{li2022neural}, shadow maps are explicitly calculated based on depth and light but are sensitive to the accumulating error in the depth map at a specific epoch\footnote{In~\cite{li2022neural}, they use the same shadow maps as \cite{li2022self} at the beginning but recalculated based on 500 epoch's normal map and remain fixed afterward.}. 
In contrast, DANI-Net overcomes these limitations by explicitly calculating shadow maps based on iteratively optimized depth and light, free from impact of surface points with dark color and accumulating errors in the depth map, leading to high-quality shadow maps.

\begin{table*}[t]
    \setlength{\abovecaptionskip}{0cm}
    \setlength{\belowcaptionskip}{0cm}
   \setlength{\tabcolsep}{12pt}
   \caption{Quantitative comparison in terms of MAE of surface normal on {\sc Ball} and {\sc Bunny} with 10 different materials from {\sc DiLiGenT}$10^2$ benchmark dataset~\cite{ren2022diligent102}. \textbf{Bold numbers} and \underline{underlined numbers} indicate the best and the second-best results, respectively.}
    \label{tab:normal_diligent100}
    \centering
    \resizebox{\linewidth}{!}{
    \begin{NiceTabular}{c|c|cccccc|ccc|c|c}
    \hline
        \multirow{2}[2]{*}{Object} & \multirow{2}[2]{*}{Method} & \multicolumn{6}{c}{Isotropic Group}           & \multicolumn{3}{c}{Anisotropic} & Chall. Group & \multirow{2}[2]{*}{AVG} \\
          &       & {\sc Pom}   & {\sc Pp}    & {\sc Nylon} & {\sc Pvc}   & {\sc Abs}   & {\sc Bakelite} & {\sc Al}    & {\sc Cu}    & {\sc Steel} & {\sc Acrylic} &  \\
        \hline
        \multirow{6}[0]{*}{{\sc Ball}} 
        & CNN-PS~\cite{ikehata2012robust} & 5.10 & 6.40 & \textbf{4.20} & 4.50 & 6.90 & 7.30 & 15.90 & \underline{14.10} & \underline{16.40} & \textbf{19.10} & \underline{9.99} \\
          & CH19~\cite{chen2019self}  & 4.40  & \textbf{2.10}  & \underline{4.60} & 4.50  & \underline{3.20} & \textbf{4.30} & \underline{15.30}  & 19.70  & 21.90  & 42.40  & 12.24  \\
          & CW20~\cite{chen2020learned}  & 10.10  & 3.01  & 9.84  & 7.13  & 5.64  & \underline{4.94}  & 21.49  & 24.01  & 27.04  & 37.92  & 15.11  \\
          & SCPS-NIR~\cite{li2022self} & \underline{3.07}  & \underline{2.30}  & 9.28  & \textbf{1.40} & 5.29  & 5.23  & 20.27  & 26.82  & 27.11  & 57.75  & 15.85  \\
          & DANI-Net  & \textbf{2.04} & 3.15  & 7.06  & \underline{4.14}  & \textbf{2.87} & 6.07  & \textbf{3.94} & \textbf{11.75} & \textbf{8.03} & \underline{34.14}  & \textbf{8.32}  \\
    \hline
    \multirow{6}[0]{*}{{\sc Bunny}} 
    & CNN-PS~\cite{ikehata2012robust} & 24.30 & 11.40 & 27.20 & 7.80 & 20.80 & \underline{9.10} & \underline{12.40} & \underline{7.70} & \underline{11.60} & \underline{14.40} & \underline{14.67} \\
          & CH19~\cite{chen2019self}  & 29.00  & 18.00  & 29.00  & 17.00  & 28.00  & 18.00  & 28.00  & 16.00  & 23.00  & 33.00  & 23.90  \\
          & CW20~\cite{chen2020learned}  & \underline{19.88}  & 11.64  & \textbf{22.36} & 9.49  & \underline{18.63}  & 12.55  & 18.03  & 12.54  & 17.85  & 28.54  & 17.15  \\
          & SCPS-NIR~\cite{li2022self} &  22.81  & \underline{9.49}  & 26.76  & \underline{7.44}  & 19.39  & 9.82  & 21.52  & 8.13  & 18.55  & 26.20  & 17.01  \\
          & DANI-Net  & \textbf{19.55}  & \textbf{7.57} & \underline{23.82}  & \textbf{6.59} & \textbf{16.49} & \textbf{7.81} & \textbf{10.36} & \textbf{6.76} & \textbf{6.16} & \textbf{21.89}  & \textbf{12.70} \\

    \hline
    \end{NiceTabular}%
    }
    \vspace{-10pt}
  \end{table*}
  
\begin{table*}[t]
\setlength{\tabcolsep}{8pt}
\setlength{\abovecaptionskip}{0cm}
\setlength{\belowcaptionskip}{0cm}
\caption{Quantitative comparison in terms of MAE of surface normal on the `anisotropic group' of {\sc DiLiGenT}$10^2$ benchmark dataset~\cite{ren2022diligent102} with 10 materials. \textbf{Bold numbers} and \underline{underlined numbers} indicate the best and the second-best results, respectively.}
    \label{tab:abl_diligent100}
    \centering
    \resizebox{\linewidth}{!}{
    \begin{tabular}{c|c|cccccccccc|c}
    \hline
     Material & Method& {\sc Ball}  & {\sc Golf}  & {\sc Spike} & {\sc Nut}   & {\sc Square} & {\sc Pentagon} & {\sc Hexagon} & {\sc Propeller} & {\sc Turbine} & {\sc Bunny} & AVG \\
    \hline
    \multirow{4}[0]{*}{{\sc Al}} 
          & CNN-PS~\cite{ikehata2018cnn} & \underline{15.90}  & 11.60  & \underline{14.30}  & \underline{16.10}  & \underline{13.40}  & \underline{14.60}  & 18.30  & \textbf{16.40}  & \textbf{25.20}  & \underline{12.40}  & \underline{15.82}  \\
          & SCPS-NIR~\cite{li2022self}  &     20.27  & 15.20  & 25.66  & 22.60  & 18.37  & 24.25  & 32.38  & 33.14  & \underline{31.20}  & 21.52  & 24.46  \\
          & DANI-Net {\it w/o} ASG & 17.51  & \underline{10.01}  & 20.45  & 17.57  & 14.22  & 14.78  & \underline{14.81}  & \underline{28.88}  & 33.61  & 12.34  & 18.42  \\
          & DANI-Net & \textbf{3.94} & \textbf{7.23} & \textbf{10.59} & \textbf{14.50} & \textbf{11.56} & \textbf{14.04} & \textbf{14.25} & 37.09  & 32.33  & \textbf{10.36} & \textbf{15.59} \\
    \hline
    \multirow{4}[0]{*}{{\sc Cu}}
          & CNN-PS~\cite{ikehata2012robust} & \underline{14.10}  & 9.20  & 8.30  & 13.00  & \textbf{4.90}  & \textbf{12.80}  & 10.40  & \textbf{9.60}  & \textbf{22.40}  & 7.70  & \underline{11.24}  \\
          & SCPS-NIR~\cite{li2022self}  & 26.82  & 9.86  & 8.92  & 25.71  & 27.16  & 39.35  & 11.89  & 14.72  & 28.96  & 8.13  & 20.15  \\
          & DANI-Net {\it w/o} ASG &     20.03  & \underline{7.07}  & \underline{7.42}  & \underline{12.03}  & 6.75  & 22.53  & \underline{7.38}  & \underline{9.75}  & 24.90  & \underline{7.40}  & 12.53  \\
          & DANI-Net & \textbf{11.75} & \textbf{6.44} & \textbf{6.08} & \textbf{10.46} & \underline{6.43}  & \underline{15.89}  & \textbf{6.66} & 10.15  & \underline{23.79}  & \textbf{6.76} & \textbf{10.44} \\
    \hline
    \multirow{4}[0]{*}{{\sc Steel}} 
          & CNN-PS~\cite{ikehata2012robust} &  \underline{16.40}  & 13.40  & \underline{16.10}  & 13.90  & \textbf{7.90}  & 15.50  & 16.90  & \textbf{9.80}  & \textbf{21.50}  & 11.60  & \underline{14.30}  \\
          & SCPS-NIR~\cite{li2022self}  &27.11  & 20.54  & 29.22  & 16.07  & 33.37  & 36.84  & 27.23  & 27.69  & 34.15  & 18.55  & 27.08  \\
          & DANI-Net {\it w/o} ASG & 20.16  & \underline{8.43}  & 22.89  & \underline{11.45}  & 17.83  & \textbf{13.60}  & \underline{13.47}  & \underline{20.88}  & \underline{24.71}  & \underline{8.92}  & 16.23  \\
          & DANI-Net & \textbf{8.03} & \textbf{7.50} & \textbf{11.85} & \textbf{9.95} & \underline{12.95}  & \underline{15.24}  & \textbf{11.29} & 21.63  & 32.66  & \textbf{6.16} & \textbf{13.73} \\
    \hline
    \end{tabular}%
    }
    \vspace{-17pt}
\end{table*}

\subsection{Performance on {\sc DiLiGenT}$10^2$~\cite{ren2022diligent102}}
To conduct an in-depth analysis of DANI-Net regarding its generalization ability on different light conditions and imaging setups, we test DANI-Net on the challenging {\sc DiLiGenT}$10^2$ benchmark dataset~\cite{ren2022diligent102}\footnote{Complete results of all 100 objects for surface normal estimation and light calibration on {\sc DiLiGenT}$10^2$ benchmark dataset~\cite{ren2022diligent102} 
can be found in the supplementary material.}.
As the supervised calibrated PS method CNN-PS~\cite{ikehata2018cnn} achieves the best performance on {\sc DiLiGenT}$10^2$~\cite{ren2022diligent102} according to~\cite{ren2022diligent102}, we also compare with it in this section.

\textbf{Surface normal estimation.}
To evaluate the performance of DANI-Net on a variety of materials, we test it on 10 different materials with two typical shapes of simple {\sc Ball} and general {\sc Bunny}.
In additional to  the state-of-the-art supervised PS method CNN-PS~\cite{ikehata2018cnn}, we also compare with three state-of-the-art UPS methods CH19~\cite{chen2019self}, CW20~\cite{chen2020learned}, and SCPS-NIR~\cite{li2022self}.
As can be observed in \Tref{tab:normal_diligent100}, 
DANI-Net achieves the best performance for surface normal estimation and even outperforms the supervised PS method CNN-PS~\cite{ikehata2018cnn}, especially for anisotropic materials of {\sc Steel}, {\sc Cu}, and {\sc Al}.
It owes to our anisotropic reflectance modeling and differentiable shadow handling method.
However, DANI-Net is less competitive on {\sc Acrylic} material with a translucent effect due to the absence of explicit modeling of translucent BRDFs. 

\textbf{Anisotropic reflectance modeling validation.}
To further testify the effectiveness of the anisotropic reflectance model in DANI-Net, we conduct experiments on the `anisotropic group' of {\sc DiLiGenT}$10^2$ benchmark dataset~\cite{shi2016benchmark} with 10 different shapes. 
In addition to the state-of-the-art methods, CNN-PS~\cite{ikehata2018cnn} (supervised and calibrated) and SCPS-NIR~\cite{li2022self}  (unsupervised and uncalibrated), we also compare with `DANI-Net {\it w/o} ASG', which is the alternative of DANI-Net with isotropic reflectance modeling. 
That is, we apply ISG bases instead of ASG bases by setting $r_k^x = r^y_k$ in \Eref{eq:asg} during training. 
As can be observed from \Tref{tab:abl_diligent100}, DANI-Net outperforms CNN-PS~\cite{ikehata2018cnn} and SCPS-NIR~\cite{li2022self} for all cases except for shapes of {\sc Propeller} and {\sc Turbine}, where the averaged MAE of these two shapes over three materials (\ie, {\sc Al}, {\sc Cu}, and {\sc Steel}) is $26.27^{\circ}$.
This is because DANI-Net takes a poor initialization of light from~\cite{li2022self} on these two shapes\footnote{As compared with~\cite{li2022self}, the GCNet in~\cite{chen2020learned} provides a better initialization of light for shapes of {\sc Propeller} and {\sc Turbine}, which improves DANI-Net's normal estimation results on these two shapes from $26.27^{\circ}$ to $16.21^{\circ}$ in average. The study of the impact on using different initialization of light can be found in the supplementary material.}.
Even so, DANI-Net achieves the best numbers in average across different shapes.
\Tref{tab:abl_diligent100} also shows an observable performance degradation of `DANI-Net {\it w/o} ASG' due to its isotropic reflectance modeling, which indicates the effectiveness of using the anisotropic reflectance model.

\section{Conclusion}
This paper proposes DANI-Net to address UPS in an unsupervised manner.
Thanks to shadow handling and anisotropic reflectance modeling, DANI-Net can address more general objects with complicated shadows and materials. 
This paper also presents solutions for shadow calculation and surface normal fitting for differentiable shadow handling. 
Our unsupervised training manner and anisotropic reflectance modeling are beneficial for generalizing DANI-Net to data from different sources.

\textbf{Limitations and future work.} Although our method produces promising results for light conditions and surfaces normal estimation on multiple real-world datasets, it has several limitations. 
First, DANI-Net cannot handle objects with strong inter-reflections or translucent materials.
Second, the image's noise caused by overexposure or underexposure will degrade the performance of the proposed DANI-Net and other inverse rendering methods. Third, we have a longer testing time compared to~\cite{li2022self} (\ie, 34 minutes for DANI-Net vs. 14 minutes for~\cite{li2022self} in average per objects in {\sc DiLiGenT}~\cite{shi2016benchmark}).
Given DANI-Net's performance on anisotropic materials, building up a more sophisticated reflectance model to consider the translucent effect and inter-reflection is worth exploring in the future.

\textbf{Acknowledgments.} This work is supported by National Natural Science Foundation of China under Grant No. 61925603, 62136001, 62088102.

\newpage
{\small
\bibliographystyle{ieee_fullname}
\bibliography{egbib}

\begin{thebibliography}{10}\itemsep=-1pt

\bibitem{ackermann2012photometric}
Jens Ackermann, Fabian Langguth, Simon Fuhrmann, and Michael Goesele.
\newblock Photometric stereo for outdoor webcams.
\newblock In {\em Proc. Computer Vision and Pattern Recognition (CVPR)}, 2012.

\bibitem{alldrin2008photometric}
Neil Alldrin, Todd Zickler, and David Kriegman.
\newblock Photometric stereo with non-parametric and spatially-varying
  reflectance.
\newblock In {\em Proc. Computer Vision and Pattern Recognition (CVPR)}, 2008.

\bibitem{barsky20034}
Svetlana Barsky and Maria Petrou.
\newblock The 4-source photometric stereo technique for three-dimensional
  surfaces in the presence of highlights and shadows.
\newblock In {\em IEEE Transactions on Pattern Analysis and Machine
  Intelligence (TPAMI)}, 2003.

\bibitem{belhumeur1999}
Peter~N Belhumeur, David~J Kriegman, and Alan~L Yuille.
\newblock The bas-relief ambiguity.
\newblock In {\em International Journal of Computer Vision (IJCV)}, 1999.

\bibitem{Boss_2021_ICCV}
Mark Boss, Raphael Braun, Varun Jampani, Jonathan~T. Barron, Ce Liu, and
  Hendrik~P.A. Lensch.
\newblock Nerd: Neural reflectance decomposition from image collections.
\newblock In {\em Proc. International Conference on Computer Vision (ICCV)},
  2021.

\bibitem{chabert2006relighting}
Charles-F{\'e}lix Chabert, Per Einarsson, Andrew Jones, Bruce Lamond, Wan-Chun
  Ma, Sebastian Sylwan, Tim Hawkins, and Paul Debevec.
\newblock Relighting human locomotion with flowed reflectance fields.
\newblock In {\em EGSR}, 2006.

\bibitem{chandraker2007shadowcuts}
Manmohan Chandraker, Sameer Agarwal, and David Kriegman.
\newblock Shadowcuts: Photometric stereo with shadows.
\newblock In {\em Proc. Computer Vision and Pattern Recognition (CVPR)}, 2007.

\bibitem{chen2019self}
Guanying Chen, Kai Han, Boxin Shi, Yasuyuki Matsushita, and Kwan-Yee~K Wong.
\newblock Self-calibrating deep photometric stereo networks.
\newblock In {\em Proc. Computer Vision and Pattern Recognition (CVPR)}, 2019.

\bibitem{chen2018ps}
Guanying Chen, Kai Han, and Kwan-Yee~K Wong.
\newblock Ps-fcn: A flexible learning framework for photometric stereo.
\newblock In {\em Proc. European Conference on Computer Vision (ECCV)}, 2018.

\bibitem{chen2020learned}
Guanying Chen, Michael Waechter, Boxin Shi, Kwan-Yee~K Wong, and Yasuyuki
  Matsushita.
\newblock What is learned in deep uncalibrated photometric stereo?
\newblock In {\em Proc. European Conference on Computer Vision (ECCV)}, 2020.

\bibitem{cho2018semi}
Donghyeon Cho, Yasuyuki Matsushita, Yu-Wing Tai, and In~So Kweon.
\newblock Semi-calibrated photometric stereo.
\newblock In {\em IEEE Transactions on Pattern Analysis and Machine
  Intelligence (TPAMI)}, 2018.

\bibitem{chung2008efficient}
Hin-Shun Chung and Jiaya Jia.
\newblock Efficient photometric stereo on glossy surfaces with wide specular
  lobes.
\newblock In {\em Proc. Computer Vision and Pattern Recognition (CVPR)}, 2008.

\bibitem{georghiades2003incorporating}
Athinodoros~S Georghiades.
\newblock Incorporating the torrance and sparrow model of reflectance in
  uncalibrated photometric stereo.
\newblock In {\em Proc. International Conference on Computer Vision (ICCV)},
  2003.

\bibitem{goldman2009shape}
Dan~B Goldman, Brian Curless, Aaron Hertzmann, and Steven~M Seitz.
\newblock Shape and spatially-varying brdfs from photometric stereo.
\newblock In {\em IEEE Transactions on Pattern Analysis and Machine
  Intelligence (TPAMI)}, 2009.

\bibitem{hashimoto2019uncalibrated}
Shuhei Hashimoto, Daisuke Miyazaki, and Shinsaku Hiura.
\newblock Uncalibrated photometric stereo constrained by intrinsic reflectance
  image and shape from silhoutte.
\newblock In {\em International Conference on Machine Vision Applications
  (MVA)}, 2019.

\bibitem{herbort2011introduction}
Steffen Herbort and Christian W{\"o}hler.
\newblock An introduction to image-based 3d surface reconstruction and a survey
  of photometric stereo methods.
\newblock {\em 3D Research}, 2011.

\bibitem{higo2010consensus}
Tomoaki Higo, Yasuyuki Matsushita, and Katsushi Ikeuchi.
\newblock Consensus photometric stereo.
\newblock In {\em Proc. Computer Vision and Pattern Recognition (CVPR)}, 2010.

\bibitem{holroyd2008photometric}
Michael Holroyd, Jason Lawrence, Greg Humphreys, and Todd Zickler.
\newblock A photometric approach for estimating normals and tangents.
\newblock In {\em ACM TOG}, 2008.

\bibitem{ikehata2018cnn}
Satoshi Ikehata.
\newblock Cnn-ps: Cnn-based photometric stereo for general non-convex surfaces.
\newblock In {\em Proceedings of the European conference on computer vision
  (ECCV)}, pages 3--18, 2018.

\bibitem{ikehata2012robust}
Satoshi Ikehata, David Wipf, Yasuyuki Matsushita, and Kiyoharu Aizawa.
\newblock Robust photometric stereo using sparse regression.
\newblock In {\em Proc. Computer Vision and Pattern Recognition (CVPR)}, 2012.

\bibitem{kajiya1986rendering}
James~T Kajiya.
\newblock The rendering equation.
\newblock In {\em Proceedings of the 13th annual conference on Computer
  graphics and interactive techniques}, pages 143--150, 1986.

\bibitem{kajiya1984ray}
James~T Kajiya and Brian~P Von~Herzen.
\newblock Ray tracing volume densities.
\newblock In {\em ACM SIGGRAPH computer graphics}, 1984.

\bibitem{karnieli2022deepshadow}
Asaf Karnieli, Ohad Fried, and Yacov Hel-Or.
\newblock Deepshadow: Neural shape from shadow.
\newblock In {\em Proc. European Conference on Computer Vision (ECCV)}, 2022.

\bibitem{kriegman2001shadows}
David~J Kriegman and Peter~N Belhumeur.
\newblock What shadows reveal about object structure.
\newblock {\em JOSA A}, 2001.

\bibitem{li2022neural}
Junxuan Li and Hongdong Li.
\newblock Neural reflectance for shape recovery with shadow handling.
\newblock In {\em Proc. Computer Vision and Pattern Recognition (CVPR)}, 2022.

\bibitem{li2022self}
Junxuan Li and Hongdong Li.
\newblock Self-calibrating photometric stereo by neural inverse rendering.
\newblock In {\em Proc. European Conference on Computer Vision (ECCV)}, 2022.

\bibitem{liu2019soft}
Shichen Liu, Tianye Li, Weikai Chen, and Hao Li.
\newblock Soft rasterizer: A differentiable renderer for image-based 3d
  reasoning.
\newblock In {\em Proc. International Conference on Computer Vision (ICCV)},
  2019.

\bibitem{lu2017symps}
Feng Lu, Xiaowu Chen, Imari Sato, and Yoichi Sato.
\newblock Symps: Brdf symmetry guided photometric stereo for shape and light
  source estimation.
\newblock In {\em IEEE Transactions on Pattern Analysis and Machine
  Intelligence (TPAMI)}, 2017.

\bibitem{mildenhall2020nerf}
Ben Mildenhall, Pratul~P Srinivasan, Matthew Tancik, Jonathan~T Barron, Ravi
  Ramamoorthi, and Ren Ng.
\newblock Nerf: Representing scenes as neural radiance fields for view
  synthesis.
\newblock In {\em Proc. European Conference on Computer Vision (ECCV)}, 2020.

\bibitem{nakagawa2015estimating}
Yosuke Nakagawa, Hideaki Uchiyama, Hajime Nagahara, and Rin-Ichiro Taniguchi.
\newblock Estimating surface normals with depth image gradients for fast and
  accurate registration.
\newblock In {\em International Conference on 3D Vision}, 2015.

\bibitem{nehab2005efficiently}
Diego Nehab, Szymon Rusinkiewicz, James Davis, and Ravi Ramamoorthi.
\newblock Efficiently combining positions and normals for precise 3d geometry.
\newblock In {\em ACM transactions on graphics (TOG)}, 2005.

\bibitem{oechsle2021unisurf}
Michael Oechsle, Songyou Peng, and Andreas Geiger.
\newblock Unisurf: Unifying neural implicit surfaces and radiance fields for
  multi-view reconstruction.
\newblock In {\em Proc. Computer Vision and Pattern Recognition (CVPR)}, 2021.

\bibitem{pagurekdifferentiable}
Dave Pagurek and Jerry Yin.
\newblock Differentiable shadow rendering, 2020.

\bibitem{papadhimitri2014closed}
Thoma Papadhimitri and Paolo Favaro.
\newblock A closed-form, consistent and robust solution to uncalibrated
  photometric stereo via local diffuse reflectance maxima.
\newblock In {\em International Journal of Computer Vision (IJCV)}, 2014.

\bibitem{qi2018geonet}
Xiaojuan Qi, Renjie Liao, Zhengzhe Liu, Raquel Urtasun, and Jiaya Jia.
\newblock Geonet: Geometric neural network for joint depth and surface normal
  estimation.
\newblock In {\em Proc. Computer Vision and Pattern Recognition (CVPR)}, 2018.

\bibitem{ren2022diligent102}
Jieji Ren, Feishi Wang, Jiahao Zhang, Qian Zheng, Mingjun Ren, and Boxin Shi.
\newblock Diligent$10^2$: A photometric stereo benchmark dataset with
  controlled shape and material variation.
\newblock In {\em Proceedings of the IEEE/CVF Conference on Computer Vision and
  Pattern Recognition}, pages 12581--12590, 2022.

\bibitem{sarno2022neural}
Francesco Sarno, Suryansh Kumar, Berk Kaya, Zhiwu Huang, Vittorio Ferrari, and
  Luc Van~Gool.
\newblock Neural architecture search for efficient uncalibrated deep
  photometric stereo.
\newblock In {\em Proc. Winter Conference on Application of Computer Vision
  (WACV)}, 2022.

\bibitem{seitz2006comparison}
Steven~M Seitz, Brian Curless, James Diebel, Daniel Scharstein, and Richard
  Szeliski.
\newblock A comparison and evaluation of multi-view stereo reconstruction
  algorithms.
\newblock In {\em Proc. Computer Vision and Pattern Recognition (CVPR)}, 2006.

\bibitem{shi2010self}
Boxin Shi, Yasuyuki Matsushita, Yichen Wei, Chao Xu, and Ping Tan.
\newblock Self-calibrating photometric stereo.
\newblock In {\em Proc. Computer Vision and Pattern Recognition (CVPR)}, 2010.

\bibitem{shi2016benchmark}
Boxin Shi, Zhe Wu, Zhipeng Mo, Dinglong Duan, Sai-Kit Yeung, and Ping Tan.
\newblock A benchmark dataset and evaluation for non-lambertian and
  uncalibrated photometric stereo.
\newblock In {\em IEEE Transactions on Pattern Analysis and Machine
  Intelligence (TPAMI)}, 2016.

\bibitem{Srinivasan_2021_CVPR}
Pratul~P. Srinivasan, Boyang Deng, Xiuming Zhang, Matthew Tancik, Ben
  Mildenhall, and Jonathan~T. Barron.
\newblock Nerv: Neural reflectance and visibility fields for relighting and
  view synthesis.
\newblock In {\em Proc. Computer Vision and Pattern Recognition (CVPR)}, 2021.

\bibitem{sunkavalli2010visibility}
Kalyan Sunkavalli, Todd Zickler, and Hanspeter Pfister.
\newblock Visibility subspaces: Uncalibrated photometric stereo with shadows.
\newblock In {\em Proc. European Conference on Computer Vision (ECCV)}, 2010.

\bibitem{taniai2018neural}
Tatsunori Taniai and Takanori Maehara.
\newblock Neural inverse rendering for general reflectance photometric stereo.
\newblock In {\em Proc. International Conference on Machine Learning (ICML)},
  2018.

\bibitem{verbin2021ref}
Dor Verbin, Peter Hedman, Ben Mildenhall, Todd Zickler, Jonathan~T Barron, and
  Pratul~P Srinivasan.
\newblock Ref-nerf: Structured view-dependent appearance for neural radiance
  fields.
\newblock In {\em Proc. Computer Vision and Pattern Recognition (CVPR)}, 2021.

\bibitem{walter2005notes}
Bruce Walter.
\newblock Notes on the ward brdf.
\newblock {\em Program of Computer Graphics, Cornell University, Technical
  Report}, 2005.

\bibitem{wang2021neus}
Peng Wang, Lingjie Liu, Yuan Liu, Christian Theobalt, Taku Komura, and Wenping
  Wang.
\newblock Neus: Learning neural implicit surfaces by volume rendering for
  multi-view reconstruction.
\newblock In {\em Proc. Conference on Neural Information Processing Systems
  (NeurIPS)}, 2021.

\bibitem{weigl2015photometric}
Eva Weigl, Sebastian Zambal, Matthias St{\"o}ger, and Christian Eitzinger.
\newblock Photometric stereo sensor for robot-assisted industrial quality
  inspection of coated composite material surfaces.
\newblock In {\em International Conference on Quality Control by Artificial
  Vision (QCAV)}, 2015.

\bibitem{woodham1980}
Robert~J Woodham.
\newblock Photometric method for determining surface orientation from multiple
  images.
\newblock In {\em Optical Engineering}, 1980.

\bibitem{wu2010robust}
Lun Wu, Arvind Ganesh, Boxin Shi, Yasuyuki Matsushita, Yongtian Wang, and Yi
  Ma.
\newblock Robust photometric stereo via low-rank matrix completion and
  recovery.
\newblock In {\em Proc. Asian Conference on Computer Vision (ACCV)}, 2010.

\bibitem{wu2006dense}
Tai-Pang Wu, Kam-Lun Tang, Chi-Keung Tang, and Tien-Tsin Wong.
\newblock Dense photometric stereo: A markov random field approach.
\newblock In {\em IEEE Transactions on Pattern Analysis and Machine
  Intelligence (TPAMI)}, 2006.

\bibitem{wu2013calibrating}
Zhe Wu and Ping Tan.
\newblock Calibrating photometric stereo by holistic reflectance symmetry
  analysis.
\newblock In {\em Proc. Computer Vision and Pattern Recognition (CVPR)}, 2013.

\bibitem{xie2015practical}
Limin Xie, Zhan Song, Guohua Jiao, Xinhan Huang, and Kui Jia.
\newblock A practical means for calibrating an led-based photometric stereo
  system.
\newblock In {\em Optics and Lasers in Engineering}, 2015.

\bibitem{xie2013real}
Wuyuan Xie, Zhan Song, and Ronald~C Chung.
\newblock Real-time three-dimensional fingerprint acquisition via a new
  photometric stereo means.
\newblock In {\em Optical Engineering}, 2013.

\bibitem{xie2022neural}
Yiheng Xie, Towaki Takikawa, Shunsuke Saito, Or Litany, Shiqin Yan, Numair
  Khan, Federico Tombari, James Tompkin, Vincent Sitzmann, and Srinath Sridhar.
\newblock Neural fields in visual computing and beyond.
\newblock In {\em Computer Graphics Forum}, 2022.

\bibitem{xu2013anisotropic}
Kun Xu, Wei-Lun Sun, Zhao Dong, Dan-Yong Zhao, Run-Dong Wu, and Shi-Min Hu.
\newblock Anisotropic spherical gaussians.
\newblock In {\em ACM TOG}, 2013.

\bibitem{yang2022ps}
Wenqi Yang, Guanying Chen, Chaofeng Chen, Zhenfang Chen, and Kwan-Yee~K Wong.
\newblock Ps-nerf: Neural inverse rendering for multi-view photometric stereo.
\newblock In {\em Proc. European Conference on Computer Vision (ECCV)}, 2022.

\bibitem{yang2022s}
Wenqi Yang, Guanying Chen, Chaofeng Chen, Zhenfang Chen, and Kwan-Yee~K Wong.
\newblock $\text{S}^3$-nerf: Neural reflectance field from shading and shadow
  under a single viewpoint.
\newblock In {\em Proc. Conference on Neural Information Processing Systems
  (NeurIPS)}, 2022.

\bibitem{yeung2014normal}
Sai-Kit Yeung, Tai-Pang Wu, Chi-Keung Tang, Tony~F Chan, and Stanley~J Osher.
\newblock Normal estimation of a transparent object using a video.
\newblock In {\em IEEE Transactions on Pattern Analysis and Machine
  Intelligence (TPAMI)}, 2014.

\bibitem{yu2002shadow}
Yizhou Yu and Johnny~T Chang.
\newblock Shadow graphs and surface reconstruction.
\newblock In {\em Proc. European Conference on Computer Vision (ECCV)}, 2002.

\bibitem{yuille1997shape}
Alan Yuille and Daniel Snow.
\newblock Shape and albedo from multiple images using integrability.
\newblock In {\em Proc. Computer Vision and Pattern Recognition (CVPR)}, 1997.

\bibitem{zhang2002novel}
Guangjun Zhang and Zhenzhong Wei.
\newblock A novel calibration approach to structured light 3d vision
  inspection.
\newblock In {\em Optics \& Laser Technology}, 2002.

\bibitem{Zhang_2021_CVPR}
Kai Zhang, Fujun Luan, Qianqian Wang, Kavita Bala, and Noah Snavely.
\newblock Physg: Inverse rendering with spherical gaussians for physics-based
  material editing and relighting.
\newblock In {\em Proc. Computer Vision and Pattern Recognition (CVPR)}, 2021.

\bibitem{zhang2021nerfactor}
Xiuming Zhang, Pratul~P Srinivasan, Boyang Deng, Paul Debevec, William~T
  Freeman, and Jonathan~T Barron.
\newblock Nerfactor: Neural factorization of shape and reflectance under an
  unknown illumination.
\newblock In {\em ACM TOG}, 2021.

\bibitem{zheng2019numerical}
Qian Zheng, Ajay Kumar, Boxin Shi, and Gang Pan.
\newblock Numerical reflectance compensation for non-lambertian photometric
  stereo.
\newblock {\em IEEE Transactions on Image Processing (TIP)}, 2019.

\bibitem{zheng2020summary}
Qian Zheng, Boxin Shi, and Gang Pan.
\newblock Summary study of data-driven photometric stereo methods.
\newblock {\em Virtual Reality \& Intelligent Hardware}, 2020.

\bibitem{zhou2010ring}
Zhenglong Zhou and Ping Tan.
\newblock Ring-light photometric stereo.
\newblock In {\em Proc. European Conference on Computer Vision (ECCV)}, 2010.

\end{thebibliography}
}

\clearpage
\newpage

\onecolumn
\section*{Supplementary Material}
\appendix

In this supplementary material,
\begin{enumerate}
    \item we provide more details about the network architectures in \Sref{sec:implement_details} (footnote 5 in the main paper);
    \item we show the complete qualitative comparison of 10 objects from {\sc DiLiGenT}~\cite{shi2016benchmark} dataset between DANI-Net and other state-of-the-art methods~\cite{chen2020learned, li2022neural, li2022self} in \Sref{sec:comp_diligent} (footnote 9 in the main paper); we also conduct additional experiments to validate the effectiveness of our normal fitting method and visualize the generated svBRDF in this section (footnote 3 in the main paper); 
    \item we conduct more ablation studies about our three-stage training schema and silhouette loss in \Sref{sec:abl} (footnote 7 in the main paper);
    \item we show the complete quantitative comparison of 100 objects (\ie, 10 shapes multiplying 10 different materials) from {\sc DiLiGenT$10^2$} dataset~\cite{ren2022diligent102} and visualization of the fitted material on 2 selected shapes (\ie, {\sc Ball}, {\sc Bunny}) with 3 anisotropic materials (\ie, {\sc Al}, {\sc Cu}, and {\sc Steel}) in \Sref{sec:comp_diligent100} (footnote 11 in the main paper); we also analyze the effects of different initialization methods in this section (footnote 12 in the main paper);
    \item we show the qualitative and quantitative results on 3 objects from {\sc Apple \& Gourd} dataset~\cite{alldrin2008photometric} and 6 objects from {\sc Light Stage Data Gallery} dataset~\cite{chabert2006relighting} in \Sref{sec:comp_apple_light} (footnote 8 in the main paper).
\end{enumerate}
The code is available at \url{https://github.com/LMozart/CVPR2023-DANI-Net}.

\newpage
\section{Implementation Details}
\label{sec:implement_details}
\textbf{Positional encoding}. We apply the same positional encoding module in NeRF~\cite{mildenhall2020nerf}\footnote{\url{https://github.com/facebookresearch/pytorch3d/tree/main/projects/nerf}} given the following equation,
\begin{equation}
\label{eq:positional encoding}
    E(p)=\left(\sin \left(2^{0} \pi p\right), \cos \left(2^{0} \pi p\right), \cdots, \sin \left(2^{L-1} \pi p\right), \cos \left(2^{L-1} \pi p\right)\right),
\end{equation}
where $L$ is the dimension of the positional code, $p$ is 2D coordinate $(u_i, v_i)$ in the image plane for DepthMLP and MaterialMLP. 
We compute the positional code $E(p)$ for each dimension and concatenate these codes to get our positional code.

\begin{figure}[h]
    \centering
    \includegraphics[width=\textwidth]{"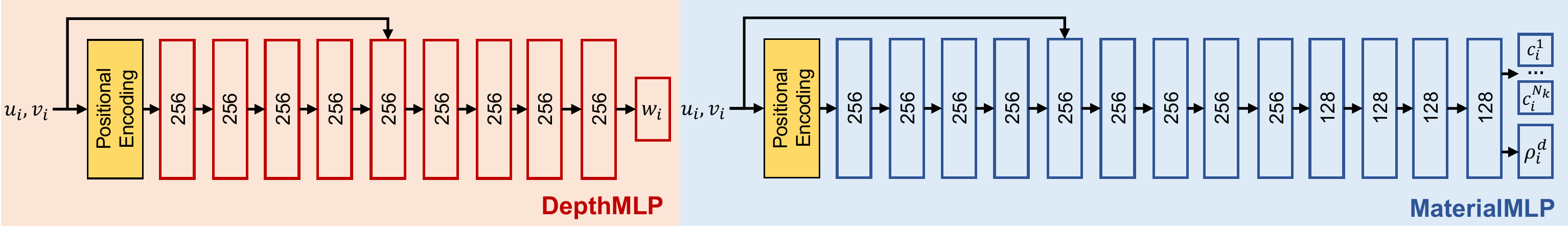"}
    \vspace{-10pt}
    \caption{Network structures of DepthMLP and MaterialMLP.} 
    \label{fig:network}
\end{figure}

\textbf{Network structure.} We show the details of the DepthMLP and MaterialMLP in \Fref{fig:network}. We use the same pre-trained light model as~\cite{li2022self} for light initialization. More details of the pre-trained light model can be found in the supplementary materials of SCPS-NIR~\cite{li2022self}.

\newpage
\section{Additional Results on {\sc DiLiGenT} Dataset~\cite{shi2016benchmark}}
\begin{figure}[h]
    \centering
    \includegraphics[width=0.8\textwidth]{"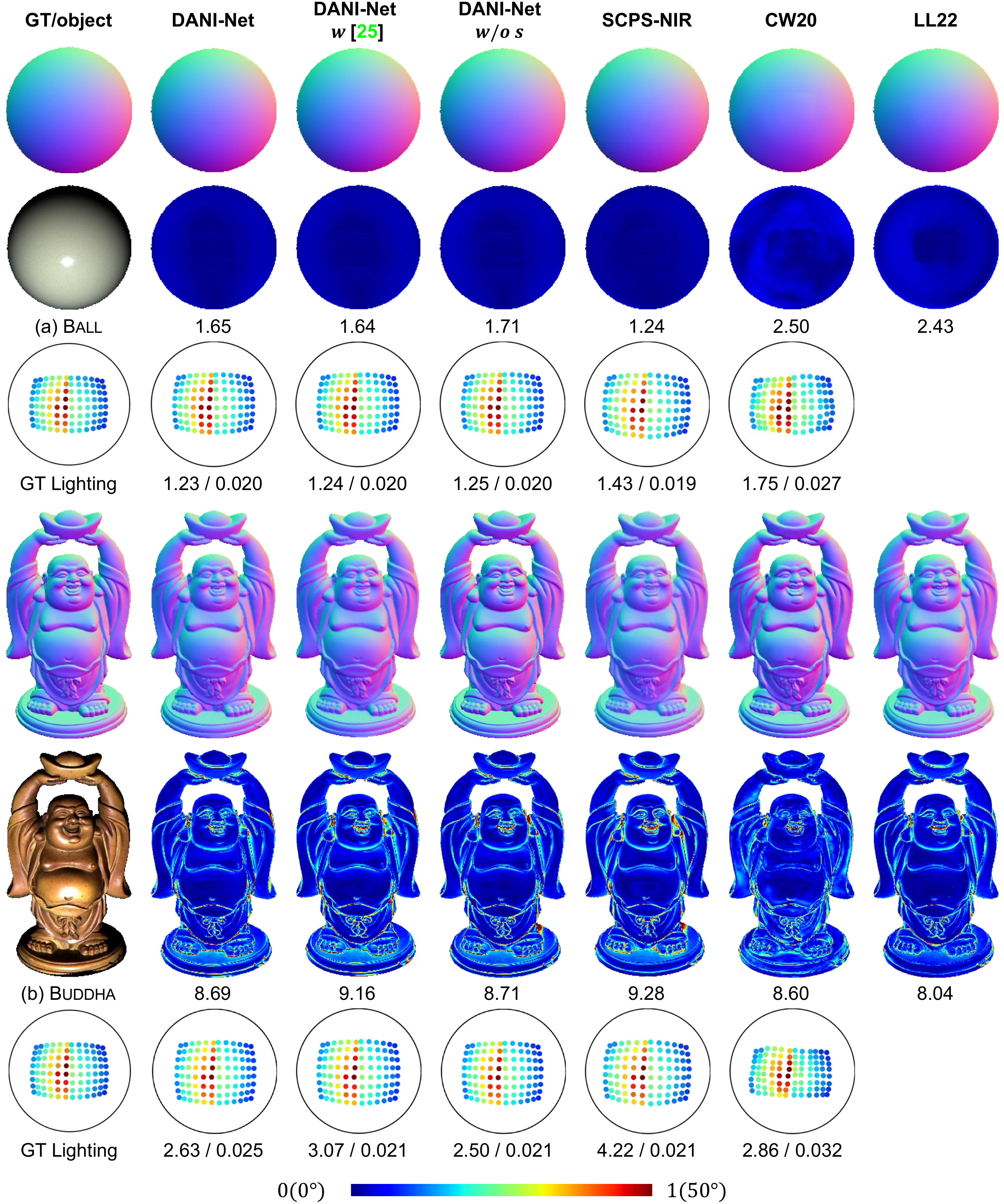"}
    \caption{The visual quality comparison among DANI-Net, DANI-Net {\it w} [\textcolor{green}{25}], DANI-Net {\it w/o} $s$, SCPS-NIR~\cite{li2022self}, CW20~\cite{chen2020learned},  and LL22~\cite{li2022neural} on {\sc Ball} and {\sc Buddha} from {\sc DiLiGenT}~\cite{shi2016benchmark} in terms of normal map (row 1,4), error map (row 2, 5), and light map (row 3, 6). Numbers indicate the MAE (for surface normal or light directions) or scale-invariant relative error (for light intensity).}
    \label{fig:diligent1}
\end{figure}

\begin{figure}[h]
    \centering
    \includegraphics[width=0.75\textwidth]{"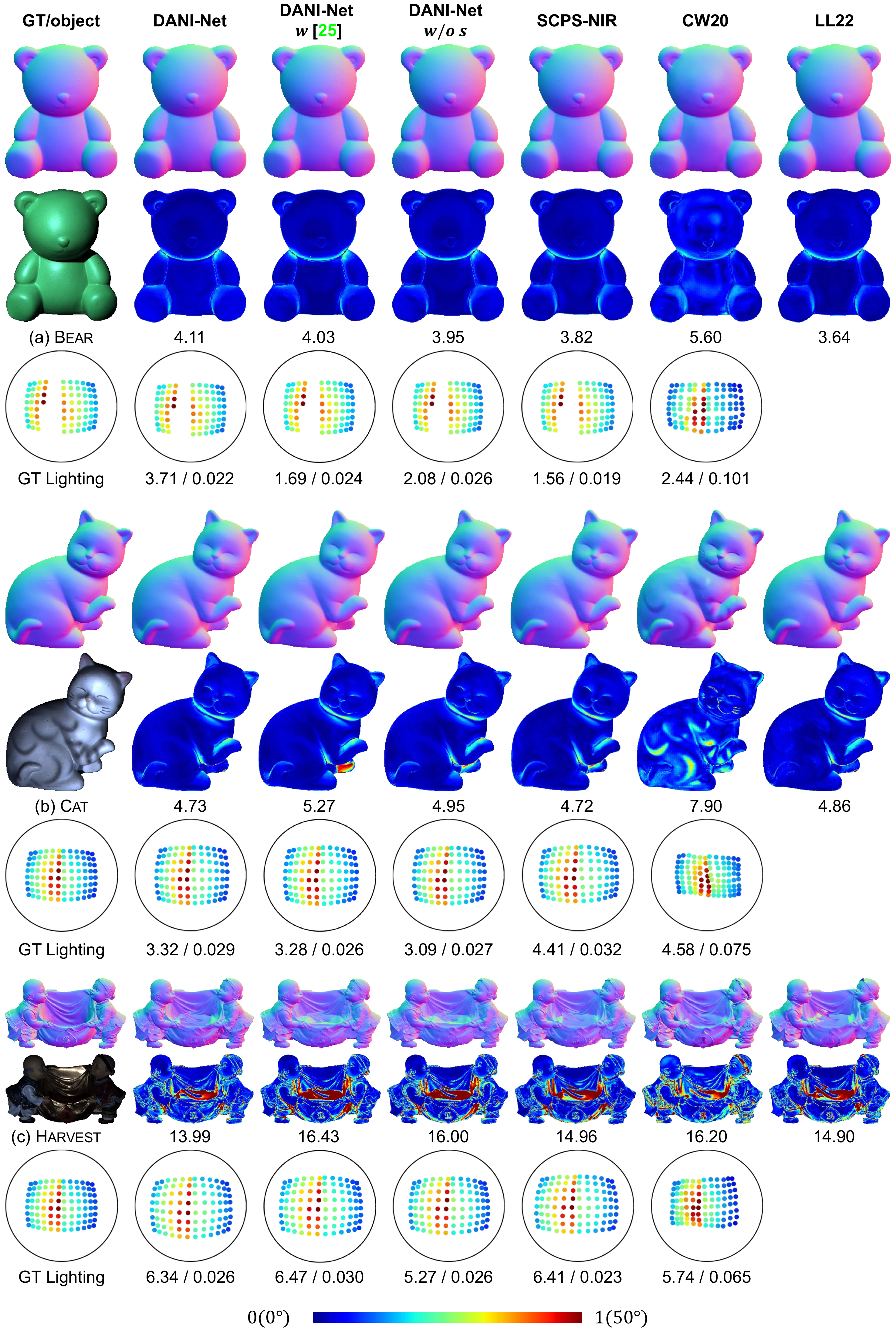"}
    \caption{The visual quality comparison among DANI-Net, DANI-Net {\it w} [\textcolor{green}{25}], DANI-Net {\it w/o} $s$, SCPS-NIR~\cite{li2022self}, CW20~\cite{chen2020learned},  and LL22~\cite{li2022neural} on {\sc Bear}, {\sc Cat}, and {\sc Harvest} from {\sc DiLiGenT}~\cite{shi2016benchmark} in terms of normal map (row 1, 4, 7), error map (row 2, 5, 8), and light map (row 3, 6, 9). Numbers indicate the MAE (for surface normal or light directions) or scale-invariant relative error (for light intensity).} 
    \label{fig:diligent2}
\end{figure}

\begin{figure}[h]
    \centering
    \includegraphics[width=0.83\textwidth]{"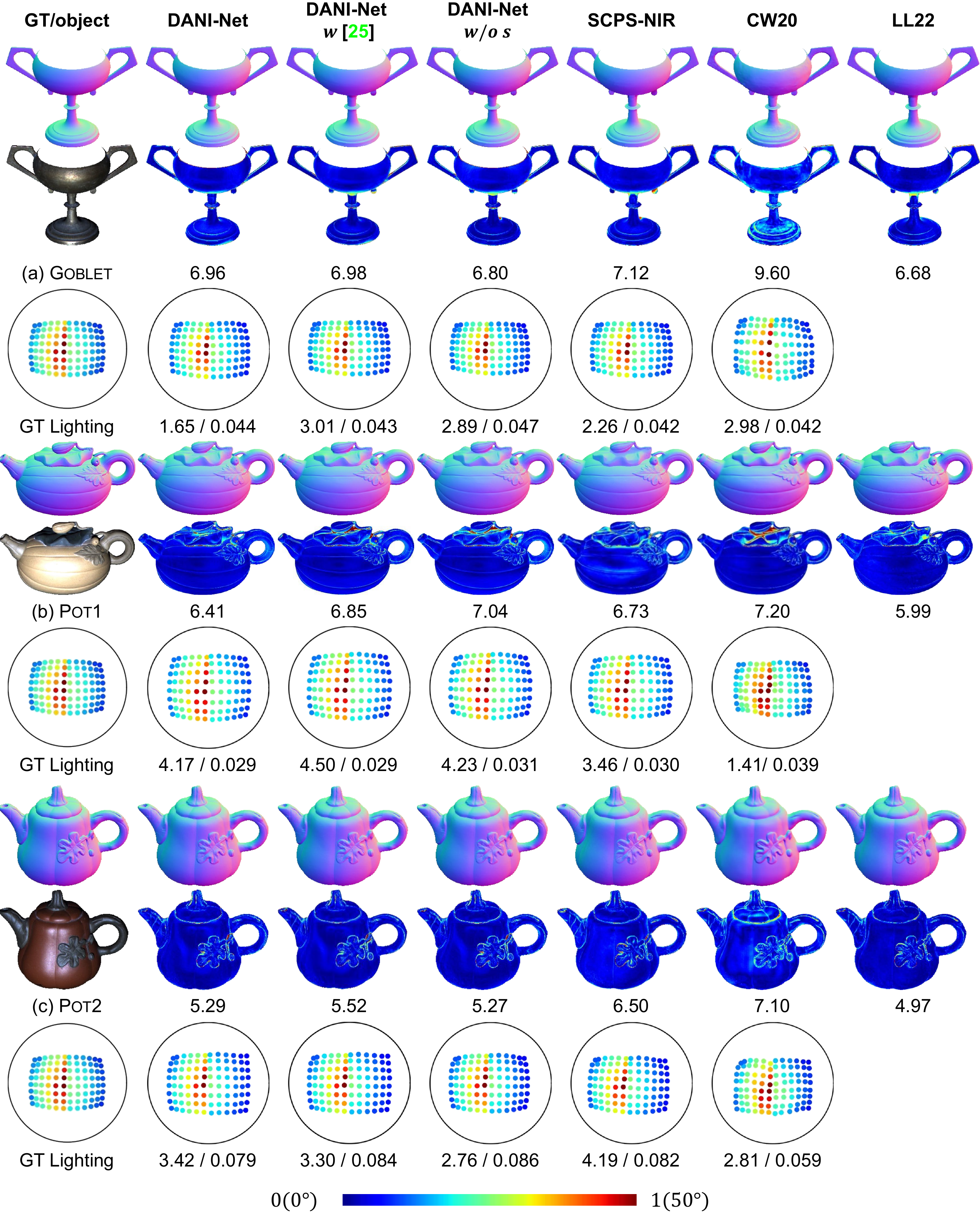"}
    \caption{The visual quality comparison among DANI-Net, DANI-Net {\it w} [\textcolor{green}{25}], DANI-Net {\it w/o} $s$, SCPS-NIR~\cite{li2022self}, CW20~\cite{chen2020learned},  and LL22~\cite{li2022neural} on  {\sc Goblet}, {\sc Pot1}, and {\sc Pot2} from {\sc DiLiGenT}~\cite{shi2016benchmark} in terms of normal map (row 1, 4, 7), error map (row 2, 5, 8), and light map (row 3, 6, 9). Numbers indicate the MAE (for surface normal or light directions) or scale-invariant relative error (for light intensity).} 
    \label{fig:diligent3}
\end{figure}

\begin{figure}[h]
    \centering
    \includegraphics[width=0.83\textwidth]{"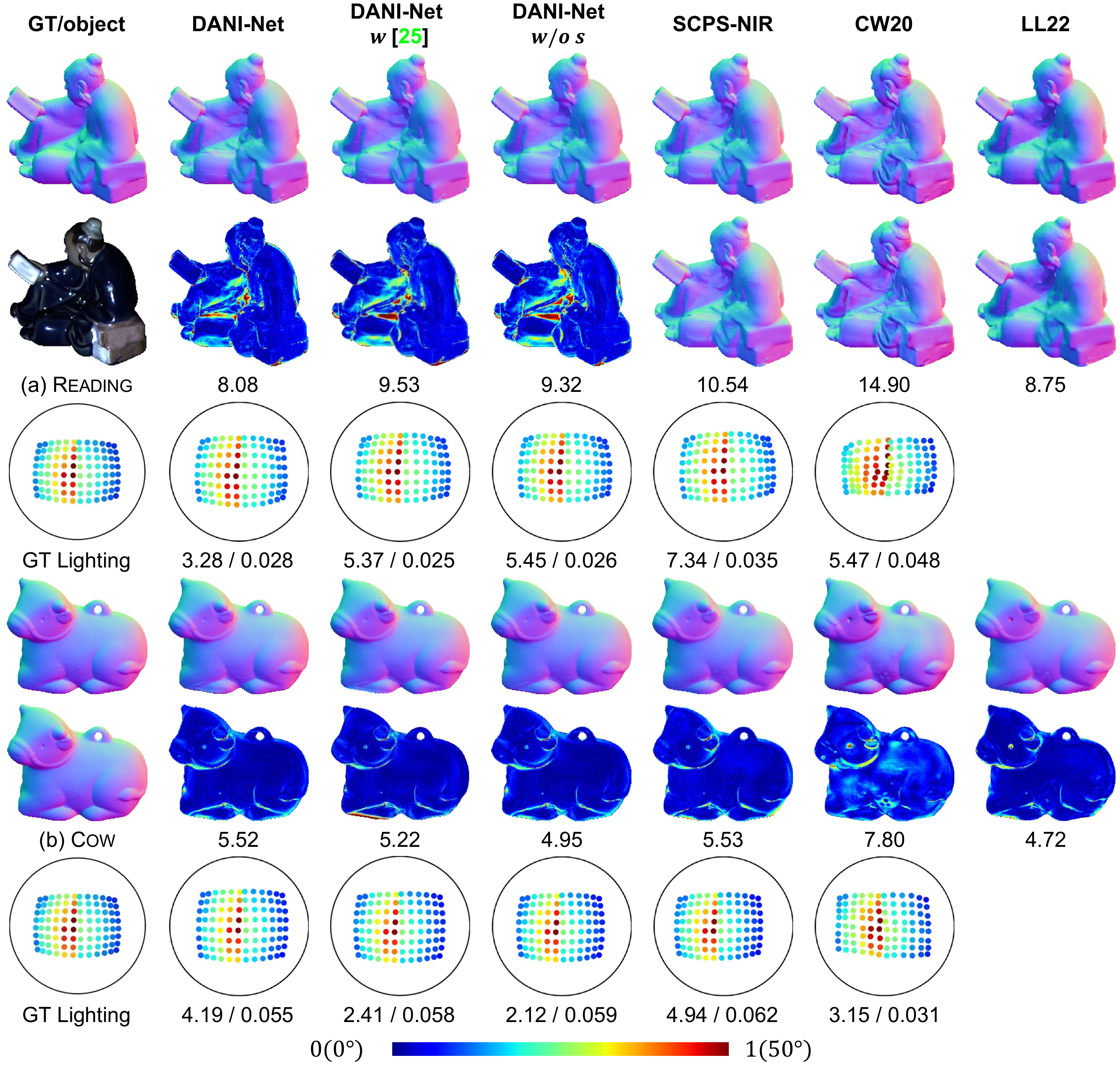"}
    \caption{The visual quality comparison among DANI-Net, DANI-Net {\it w} [\textcolor{green}{25}], DANI-Net {\it w/o} $s$, SCPS-NIR~\cite{li2022self}, CW20~\cite{chen2020learned},  and LL22~\cite{li2022neural} on  {\sc Reading} and {\sc Cow} from {\sc DiLiGenT}~\cite{shi2016benchmark} in terms of normal map (row 1,4), error map (row 2, 5), and light map (row 3, 6). Numbers indicate the MAE (for surface normal or light directions) or scale-invariant relative error (for light intensity).} 
    \label{fig:diligent4}
\end{figure}

\textbf{Qualitative results on {\sc DiLiGenT} dataset~\cite{shi2016benchmark}.}
From \Fref{fig:diligent1} to \Fref{fig:diligent4} we show the normal map, error map, and light map of 10 objects from {\sc DiLiGenT} Dataset~\cite{shi2016benchmark} predicted by DANI-Net and its alternatives (\ie, DANI-Net {\it w} [\textcolor{green}{25}] and DANI-Net {\it w/o} $s$). We further compare DANI-Net with recent UPS methods~\cite{li2022self, chen2020learned} and state-of-the-art unsupervised PS methods~\cite{li2022neural}. The results illustrate the effectiveness of the differentiable shadow handling and anisotropic material modeling method in DANI-Net.
\label{sec:comp_diligent}
\newpage
\vspace*{\fill}
\vspace*{\fill}\clearpage

\newpage
\vspace*{\fill}
\vspace*{\fill}\clearpage

\newpage
\vspace*{\fill}
\vspace*{\fill}\clearpage

\newpage
\textbf{Validation of weighted interpolation.}
To validate the effectiveness of our weighted interpolation for normal fitting, we compare DANI-Net with `DANI-Net {\it w} sobel' (\ie, Sobel operator), `DANI-Net {\it w} cross' (\ie, method in~\cite{li2022neural}), and `DANI-Net {\it w} triangle' (\ie, method in~\cite{nehab2005efficiently}). Those alternatives are only different in the normal fitting method. As shown in \Tref{tab:comp_diligent}, DANI-Net outperforms all alternatives on average, especially on objects like {\sc Harvest} with a complicated shape. We observe a similar performance between `DANI-Net {\it w} sobel' and `DANI-Net {\it w} cross', whose average MAE on normal estimation increase $0.15^{\circ}$ and $0.09^{\circ}$, respectively since they either ignore query point's information or using excessive neighbor points. The average MAE of `DANI-Net {\it w} triangle' increase about $0.39^{\circ}$ due to its ineffectiveness in backpropagation (\ie, limit points affected by the gradients).
\begin{table*}[h]
    \setlength{\tabcolsep}{10pt}
\caption{Quantitative comparison in terms of MAE of surface normal  on {\sc DiLiGenT} benchmark dataset~\cite{shi2016benchmark} for different normal fitting methods. \textbf{Bold numbers} indicates the best results.}
\vspace{-10pt}
    \label{tab:comp_diligent}
    \centering
    \resizebox{1\linewidth}{!}{
    \begin{tabular}{c|cccccccccc|c}
    \hline
    Method & \sc{Ball}  & \sc{Bear}  & \sc{Buddha} & \sc{Cat}   & \sc{Cow}   & \sc{Goblet} & \sc{Harvest} & \sc{Pot1}  & \sc{Pot2}  & \sc{Reading} & AVG    \\ 
    \hline
    DANI-Net {\it w} triangle & 2.33  & 4.27  & 9.19  & 4.87  & 5.65  & 7.44  & 15.22 & 6.61  & 5.69  & 8.01  & 6.93  \\
    DANI-Net {\it w} sobel & 1.89  & \textbf{3.97}  & 9.05  & 4.71  & \textbf{5.31}  & 7.27  & 14.85  & 6.72  & 5.42  & 7.68  & 6.69  \\
    DANI-Net {\it w} cross & 1.88  & 3.98  & 9.08  & \textbf{4.70}  & 5.36  & \textbf{6.67}  & 14.93  & 6.63  & 5.43  & \textbf{7.68}  & 6.63 \\
    DANI-Net & \textbf{1.65}   & 4.11  & \textbf{8.69}  & 4.73  & 5.52  & 6.96  & \textbf{13.99} & \textbf{6.41}  & \textbf{5.29}  & 8.08  & \textbf{6.54} \\
    \hline
    \end{tabular}
    }
    \vspace{-10pt}
\end{table*}

\begin{figure}[h]
    \centering
    \includegraphics[width=0.8\textwidth]{"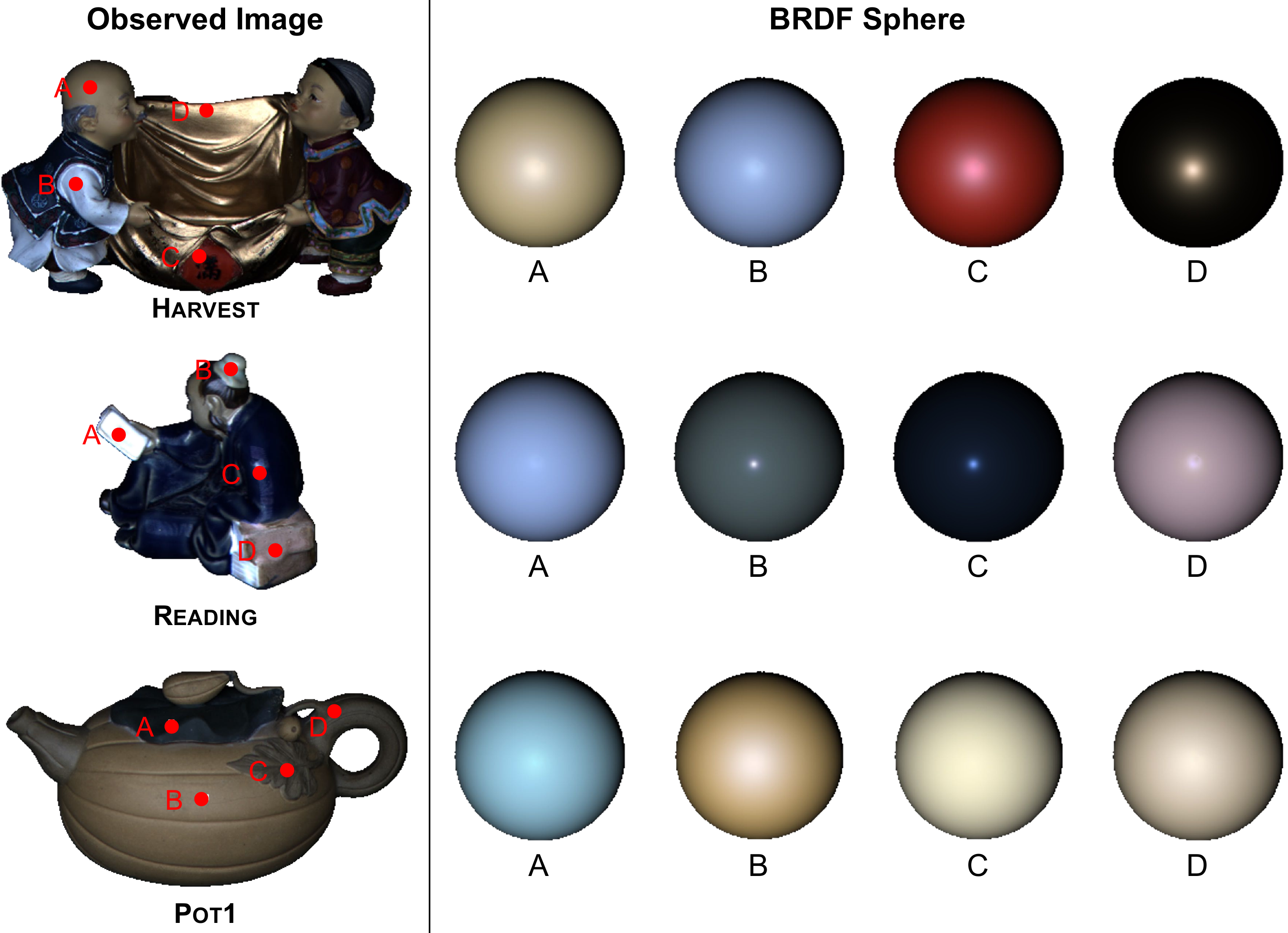"}
    \caption{Visualization of the estimated svBRDF on {\sc Harvest}, {\sc Reading}, and {\sc Pot1} from DiLiGenT dataset~\cite{shi2016benchmark}. On each object, we select four points (red dots on the first column's images, labeled ABCD) at different positions with different materials to showcase the predicted svBRDF spheres. The overall intensity of the observed images and BRDF spheres is scaled for better visualization.} 
    \label{fig:svbrdf}
\end{figure}

\textbf{Validation of svBRDF.} To validate the spatially varying BRDF (svBRDF) generated by DANI-Net, we visualize the BRDF spheres at different points of {\sc Harvest}, {\sc Reading}, and {\sc Pot1} in \Fref{fig:svbrdf}. The spheres show polychrome appearances with different roughness, demonstrating the visually pleasant svBRDF generated by DANI-Net.

\clearpage
\newpage
\section{Additional Ablation Studies}
\label{sec:abl}
\subsection{Three-stage Training Schema}
The smoothness terms incorporated in the three-stage training schema (\ie, $\mathcal{L}_{R^d}, \mathcal{L}_W,$ and $\mathcal{L}_N$) facilitate DANI-Net's convergence at an early stage (stage 1); while they are gradually dropped in the subsequent stage (stage 2, 3) to avoid over-smooth results. To validate the effectiveness of the three-stage training schema, we compare DANI-Net with: 1) `DANI-Net {\it w} $\mathcal{L}_N$' that keeps $\mathcal{L}_N$ till the training complete; 2) `DANI-Net {\it w/o} $\mathcal{L}_{R^d}$' that drops material smoothness term (\ie, $\mathcal{L}_{R^d}$) at the beginning on {\sc DiLiGenT} dataset~\cite{shi2016benchmark}; 3) `DANI-Net {\it w/o} $\mathcal{L}_N + \mathcal{L}_W$' that drops geometry smoothness terms (\ie, $\mathcal{L}_N$ and $\mathcal{L}_W$) at the beginning. As shown in \Tref{tab:3-stage}, the average MAE of normal estimation on 10 objects increases $0.22^{\circ}$, $0.28^{\circ}$, and $0.10^{\circ}$ for `DANI-Net {\it w} $\mathcal{L}_N$', `DANI-Net {\it w/o} $\mathcal{L}_{R^d}$', and `DANI-Net {\it w/o} $\mathcal{L}_{N} + \mathcal{L}_W$', respectively, indicating the necessity of the three-stage training schema. 

\begin{table*}[h]
    \setlength{\tabcolsep}{10pt}
\caption{Quantitative comparison in terms of MAE of surface normal  on {\sc DiLiGenT} benchmark dataset~\cite{shi2016benchmark} for different alternatives on three-stage training schema. \textbf{Bold numbers} indicates the best results.}
\vspace{-10pt}
    \label{tab:3-stage}
    \centering
    \resizebox{1\linewidth}{!}{
    \begin{tabular}{c|cccccccccc|c}
    \hline
    Method & \sc{Ball}  & \sc{Bear}  & \sc{Buddha} & \sc{Cat}   & \sc{Cow}   & \sc{Goblet} & \sc{Harvest} & \sc{Pot1}  & \sc{Pot2}  & \sc{Reading} & AVG    \\ 
    \hline
    DANI-Net {\it w} $\mathcal{L}_N$  & 1.57  & 4.10  & 9.76  & 4.99  & \textbf{5.16} & 6.96  & 14.41  & 7.28  & 5.88  & 7.47  & 6.76  \\
    DANI-Net {\it w/o} $\mathcal{L}_{R^d}$ & 1.47  & \textbf{4.05} & 8.79  & 5.15  & 5.69  & 6.97  & 14.28  & 7.13  & \textbf{5.24} & 9.43  & 6.82  \\
    DANI-Net {\it w/o} $\mathcal{L}_N+\mathcal{L}_W$ & \textbf{1.44} & 4.28  & \textbf{8.19} & 4.98  & 5.92  & \textbf{6.88} & 15.29  & \textbf{6.08} & 5.28  & \textbf{8.06} & 6.64  \\
    DANI-Net & 1.65  & 4.11  & 8.69  & \textbf{4.73} & 5.52  & 6.96  & \textbf{13.99} & 6.41  & 5.29  & 8.08  & \textbf{6.54} \\
    \hline
    \end{tabular}
    }
\end{table*}

\clearpage
\newpage
\subsection{Silhouette Normal}
Although the fitted silhouette normal is only reliable at the occluding boundaries, the silhouette loss $\mathcal{L}_{\text{Si}}$ is crucial in alleviating GBR ambiguity that facilitates DANI-Net's convergence and prevents it from falling into the local optimum. We compare DANI-Net with `DANI-Net {\it w/o} $\mathcal{L}_{\text{Si}}$' (or $\mathcal{L}_{\text{Si}}$) that drops $\mathcal{L}_{\text{Si}}$ at the beginning (or at stage 3) on {\sc DiLiGenT} dataset~\cite{shi2016benchmark}. As shown in \Tref{tab:silhouette}, the average MAE of normal estimation on 10 objects increases $0.52^{\circ}$ (or $0.14^{\circ}$), indicating the necessity of silhouette loss. 

\begin{figure}[h]
    \centering
    \includegraphics[width=0.6\textwidth]{"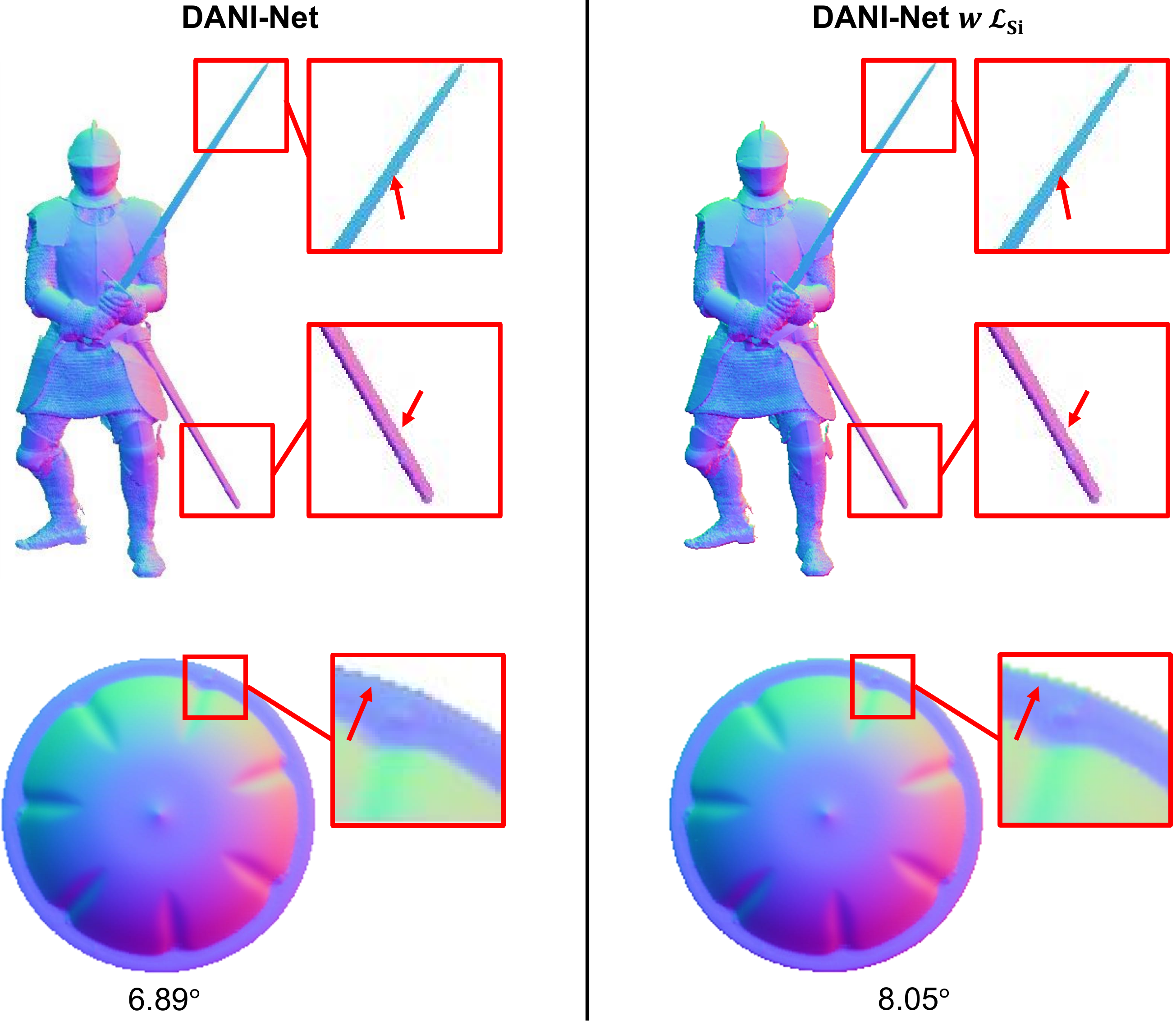"}
    \caption{Visualization of the estimated normal map on {\sc Knight Fighting} (top), and {\sc Nut Pom} (bottom) from {\sc Light Stage Data Gallery}~\cite{chabert2006relighting} and {\sc DiLiGenT}$10^2$~\cite{ren2022diligent102}. Sub-figures in the left (or right) column are normal maps estimated by DANI-Net (or `DANI-Net {\it w} $\mathcal{L}_\text{Si}$') Red boxes highlight prominent regions for easy comparison. Red arrows point to the silhouette of objects. Numbers at the bottom indicate the MAE of normal estimation.} 
    \label{fig:silh}
\end{figure}

For objects in {\sc Light Stage Data Gallery}~\cite{chabert2006relighting} and {\sc DiLiGenT} $10^2$~\cite{ren2022diligent102} with non-occluding silhouette, DANI-Net uses silhouette loss in a more flexible way. That is, we either drop the silhouette loss at stage 3 ({\sc Knight Kneeling}, {\sc Knight Fighting}, and {\sc Knight Standing} from {\sc Light Stage Data Gallery}~\cite{chabert2006relighting}), or roughly estimate the silhouette's normal instead of pre-computing it through silhouette's perpendicular vectors ({\sc Bunny, Hexagon, Nut, Pentagon, Propeller, Square,} and {\sc Turbine} from {\sc DiLiGenT} $10^2$~\cite{ren2022diligent102})\footnote{It is easy to estimate the silhouette normal of those objects as $[0, 0, 1]$. We apply the same measurement in SCPS-NIR~\cite{li2022self} to calculate the silhouette normal loss for a fair comparison.}. To clarify the necessity of this strategy, we visualize the estimated normal map of DANI-Net and `DANI-Net {\it w} $\mathcal{L}_\text{Si}$' that keeps $\mathcal{L}_\text{Si}$ in all stages on {\sc Knight Fighting} with partially non-occluding silhouettes on the sword (the top row of \Fref{fig:silh}), and {\sc Nut Pp} with predominantly non-occluding silhouettes on its base (the bottom row of \Fref{fig:silh}). It can be observed that keeping $\mathcal{L}_{\text{Si}}$ in all stages does not significantly affect the normal estimation results for objects with partially occluding silhouettes like {\sc Knight Fighting}, but it does harm the performance for objects with predominantly non-occluding silhouettes (MAE on normal estimation increases $1.16^{\circ}$ on {\sc Nut Pp}). This validates the effectiveness of our flexible strategy using $\mathcal{L}_{\text{Si}}$ on objects with non-occluding silhouettes.

\begin{table*}[h]
    \setlength{\tabcolsep}{10pt}
\caption{Quantitative comparison in terms of MAE of surface normal  on {\sc DiLiGenT} benchmark dataset~\cite{shi2016benchmark} for different alternatives on silhouette loss. \textbf{Bold numbers} indicates the best results.}
\vspace{-10pt}
    \label{tab:silhouette}
    \centering
    \resizebox{1\linewidth}{!}{
    \begin{tabular}{c|cccccccccc|c}
    \hline
    Method & \sc{Ball}  & \sc{Bear}  & \sc{Buddha} & \sc{Cat}   & \sc{Cow}   & \sc{Goblet} & \sc{Harvest} & \sc{Pot1}  & \sc{Pot2}  & \sc{Reading} & AVG    \\ 
    \hline
    DANI-Net {\it w/o} $\mathcal{L}_{\text{Si}}$ & 1.68  & 4.22  & 9.44  & 5.30  & 5.79  & 8.12  & 14.18  & 7.43  & 5.53  & 8.89  & 7.06  \\
    DANI-Net {\it w/o} $\mathcal{L}_{\text{Si}}$-3 & 1.81  & 4.07  & 8.72  & 5.15  & 5.69  & 7.10  & 13.77  & 6.95  & 5.26  & 8.30  & 6.68  \\
    DANI-Net & \textbf{1.65} & \textbf{4.11} & \textbf{8.69} & \textbf{4.73} & \textbf{5.52} & \textbf{6.96} & \textbf{13.99} & \textbf{6.41} & \textbf{5.29} & \textbf{8.08} & \textbf{6.54} \\
    \hline
    \end{tabular}
    }
    \vspace{-10pt}
\end{table*}

\clearpage
\newpage
\section{Additional Results on {\sc DiLiGenT$10^2$} Dataset~\cite{ren2022diligent102}}
\label{sec:comp_diligent100}
\begin{figure}[h]
    \centering
    \includegraphics[width=\textwidth]{"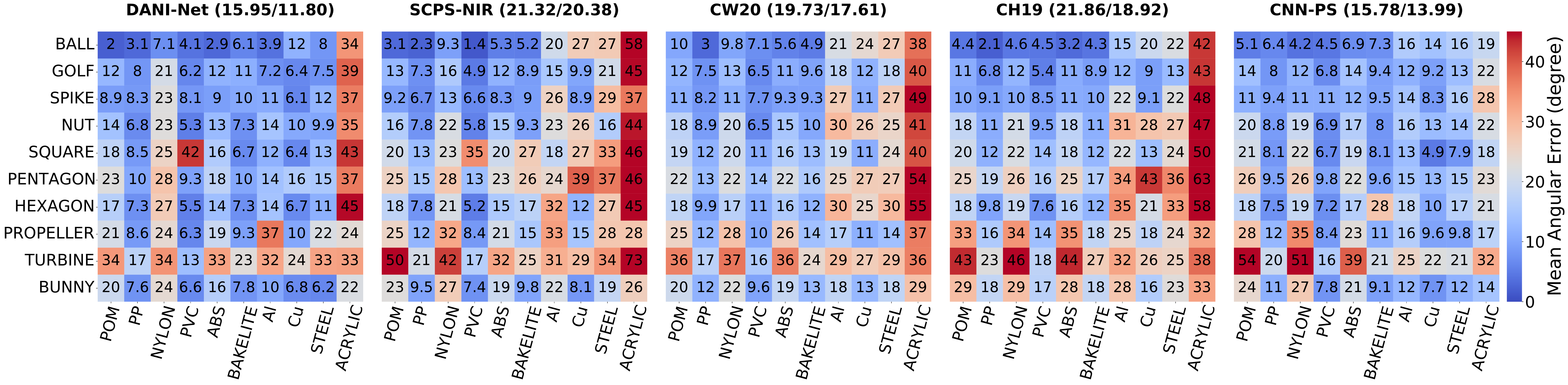"}
    \caption{Normal estimation results on {\sc DiLiGenT$10^2$} Dataset~\cite{ren2022diligent102} visualized by Shape-material error matrix~\cite{ren2022diligent102}. Numbers in the bracket indicate each method's mean/median MAE for normal estimation.} 
    \label{fig:diligent100_nor}
    \vspace{-10pt}
\end{figure}

\textbf{Quantitative results of normal estimation.} \Fref{fig:diligent100_nor} presents the comprehensive normal estimation results compared with CW20~\cite{chen2020learned}, SCPS-NIR~\cite{li2022self}, and CNN-PS~\cite{ikehata2018cnn}. While DANI-Net outperforms current methods on {\sc Ball} and {\sc Bunny} or anisotropic group (\ie, {\sc Al}, {\sc Cu}, and {\sc Steel}), the average MAE of DANI-Net on 100 objects' normal estimation is slightly higher than CNN-PS (still lower than current UPS methods). Particularly, we find that DANI-Net is not competitive on objects of {\sc Nylon} and {\sc Acrylic}, and objects like {\sc Turbine} and {\sc Propeller} with complicated shapes in the `anisotropic group'.

\begin{figure}[h]
    \vspace{-5pt}
    \centering
    \includegraphics[width=0.80\textwidth]{"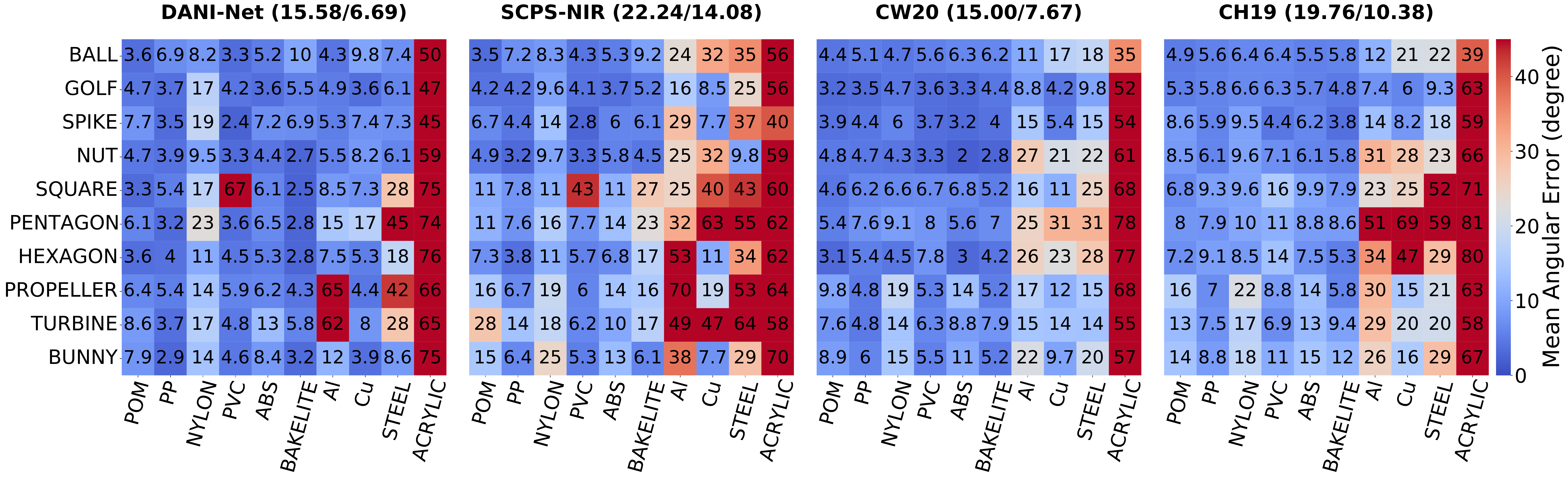"}
    \caption{Light direction calibration results on {\sc DiLiGenT$10^2$} Dataset~\cite{ren2022diligent102} visualized by Shape-material error matrix~\cite{ren2022diligent102}. Numbers in the bracket indicate each method's mean/median MAE for light direction calibration.} 
    \label{fig:diligent100_dir}
    \vspace{-10pt}
\end{figure}

\begin{figure}[h]
    \vspace{-5pt}
    \centering
    \includegraphics[width=0.80\textwidth]{"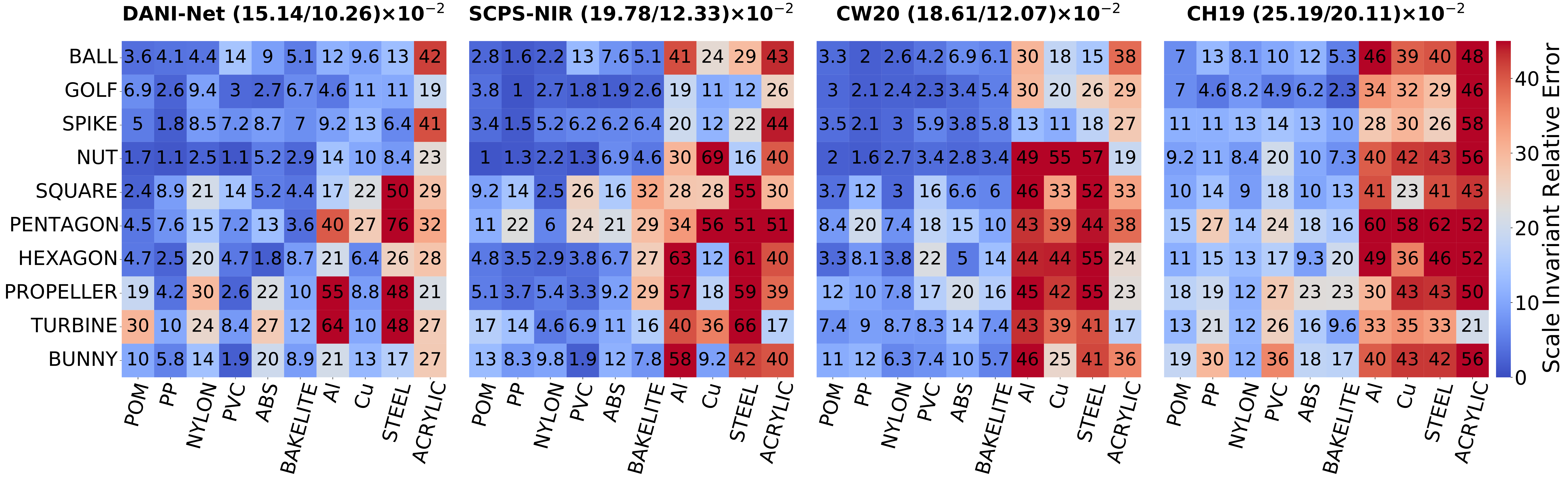"}
    \caption{Light intensity calibration results on {\sc DiLiGenT$10^2$} Dataset~\cite{ren2022diligent102} visualized by Shape-material error matrix~\cite{ren2022diligent102}. The number in the matrix is scaled for visualization. Numbers in the bracket indicate each method's mean/median scale invariant relative error for light intensity calibration.} 
    \vspace{-10pt}
    \label{fig:diligent100_int}
\end{figure}

\textbf{Quantitative results of light calibration.} \Fref{fig:diligent100_dir} and \Fref{fig:diligent100_int} present the light calibration results, compared with CW20~\cite{chen2020learned}, SCPS~\cite{chen2019self}, and SCPS-NIR~\cite{li2022self}. As can be observed in those figures, DANI-Net reaches the second-best performance on light direction calibration and the best on light intensity calibration. By cross-comparing the results of normal estimation, we find that the cases with high normal estimation MAE also result in particularly high MAE in light calibration. 

\begin{figure}[h]
    \centering
    \includegraphics[width=0.5\textwidth]{"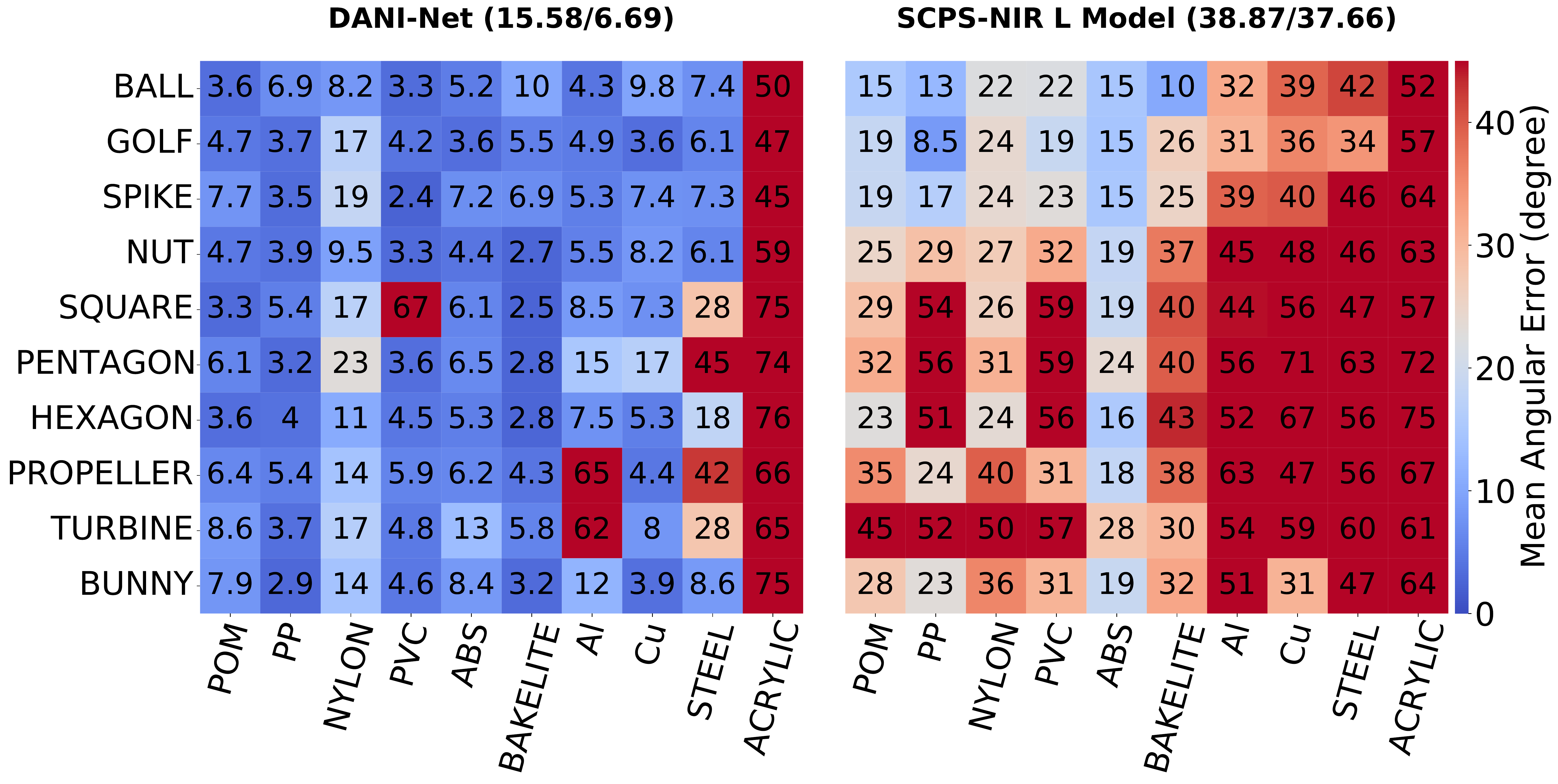"}
    \caption{Comparison between DANI-Net and the pre-trained light model of SCPS-NIR~\cite{li2022self} on light calibration. Numbers in the bracket indicate each method's mean/median MAE for light direction calibration.} 
    \label{fig:comp0}
    \vspace{-10pt}
\end{figure}

\clearpage
\newpage
\textbf{Analysis on light initialization.} We first conjecture that inadequate light initialization is the main cause of inaccurate normal estimation. To validate that, we showcase the initial light's (calibrated by the pre-trained light model~\cite{li2022self}) MAE in \Fref{fig:comp0}.

\begin{figure}[h]
    \centering
    \includegraphics[width=0.85\textwidth]{"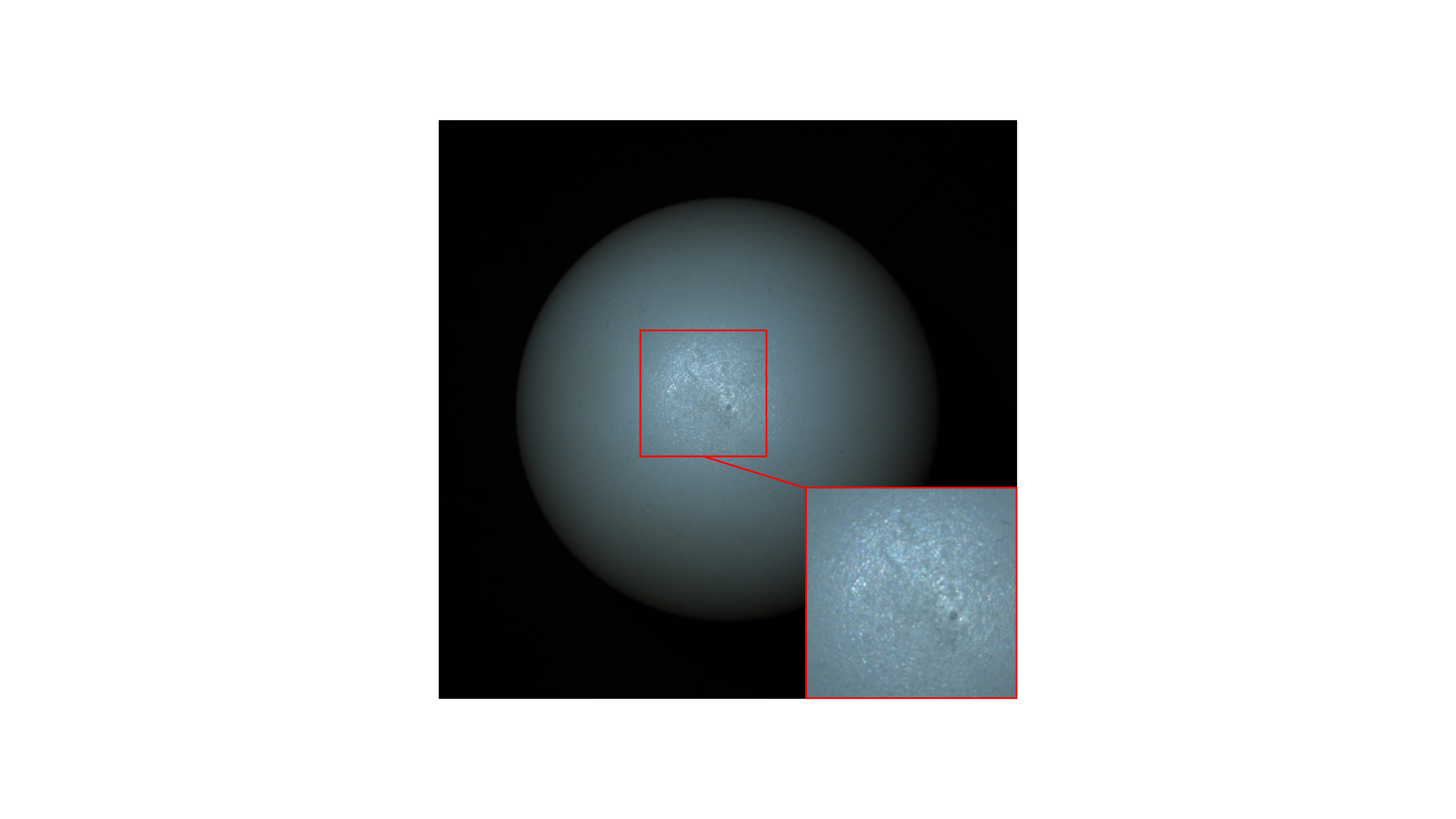"}
    \caption{Comparison between DANI-Net and DANI-Net+GCNet on normal estimation and light calibration. All of the results are visualized by the Shape-material error matrix~\cite{ren2022diligent102}. Numbers in the top and bottom left brackets indicate each method's mean/median MAE for light direction calibration. Numbers in the bottom right bracket indicate each method's mean/median scale invariant relative error for light intensity calibration.} 
    \label{fig:comp1}
    \vspace{-10pt}
\end{figure}

The results validate our conjecture, \ie, the initial lights deviate significantly from the ground truth in the aforementioned cases. Therefore, initializing the light more accurately will be a reasonable strategy (\eg, light calibrated by GCNet~\cite{chen2020learned}). We compare DANI-Net with `DANI-Net+GCNet', with the latter differing only in its light initialized by GCNet. As shown in~\Fref{fig:comp1}, the average MAE of normal estimation (or light calibration) of `DANI-Net+GCNet' decreases $0.77^\circ$ (or $3.06^\circ$). Particularly, we observe a compelling enhancement for objects with complicated shapes (\ie, {\sc Propeller}, {\sc Turbine}), but a marginal improvement for objects made of {\sc Nylon} and {\sc Acrylic}. While objects made of {\sc Acrylic} still have inaccurate light initialization leading to unsatisfactory light calibration and normal estimation\footnote{This may mainly be caused by the sub-surface scattering of translucent materials.}, the results obtained for objects made of {\sc Nylon} contradict our earlier conjecture. We further assume that the materials' exceptional properties may be the primary cause of the subpar performance.

\begin{figure}[h]
    \centering
    \includegraphics[width=0.5\textwidth]{"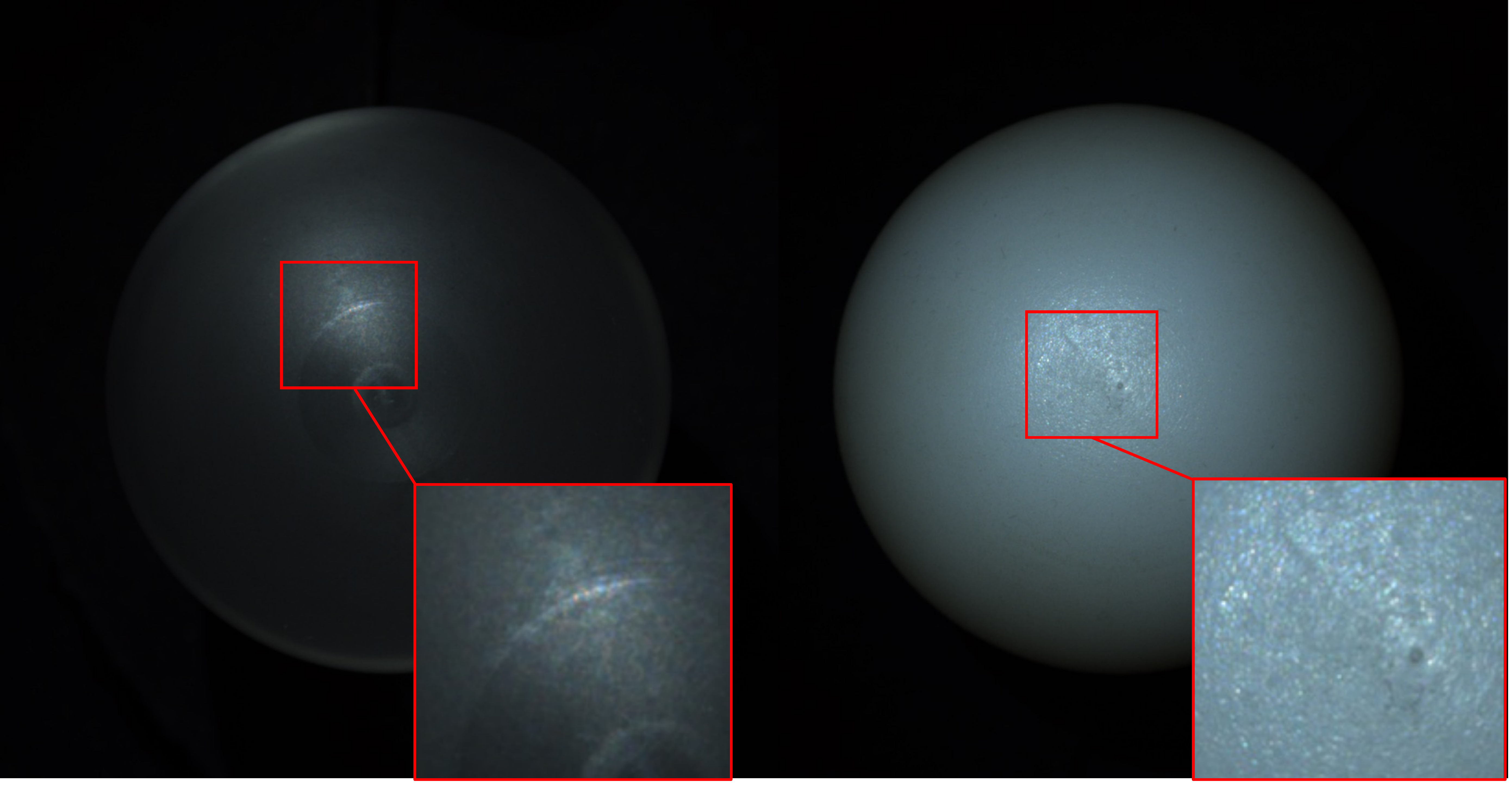"}
    \caption{The observed image of {\sc Acrylic Ball} (left) and {\sc Nylon Ball} (right) from {\sc DiLiGenT$10^2$} Dataset~\cite{ren2022diligent102}. The red squares indicate the regions that may harm the performance of DANI-Net.} 
    \label{fig:comp2}
\end{figure}

\clearpage
\newpage
\textbf{Analysis on material properties.} 
We check the observed images of {\sc Nylon Ball} (shown in the right of \Fref{fig:comp2}) and find that some images are contaminated by unexpected artifacts (see the red square in the right sub-figure) regarded as the noise that harms DANI-Net's performance. A naive trick for noise removal is to exclude the pixel points with too high or too low intensity for a smooth intensity profile. However, since the specularity contains solid reflectance cues to solve UPS, we only remove the low-intensity pixel points (\ie, the point with an intensity that is lower than the 25th percentile of overall pixels' intensities). We denote DANI-Net implementing this trick as `DANI-Net {\it w rm}'. According to \Tref{tab:comp2}, we find this simple trick further improves DANI-Net's performance on {\sc Nylon} and even remains effective on {\sc Acrylic} objects because the sub-surface scattering of translucent materials can also be regarded as noise, shown in the left of \Fref{fig:comp2}. The MAE of the surface normal for {\sc Nylon} (or {\sc Acrylic}) objects decreases $5.18^{\circ}$ (or $2.24^{\circ}$). However, note that this trick cannot explicitly exclude the noise, we regard a robust method of modeling these materials as part of our future work.

\begin{table*}[h]
\setlength{\tabcolsep}{8pt}
\caption{Quantitative comparison in terms of MAE of surface normal on {\sc DiLiGenT}$10^2$ benchmark dataset~\cite{shi2016benchmark}. \textbf{Bold numbers} indicates the best results.}
    \label{tab:comp2}
    \centering
    \resizebox{\linewidth}{!}{
    \begin{tabular}{c|c|cccccccccc|c}
    \hline
     Material & Method& {\sc Ball}  & {\sc Golf}  & {\sc Spike} & {\sc Nut}   & {\sc Square} & {\sc Pentagon} & {\sc Hexagon} & {\sc Propeller} & {\sc Turbine} & {\sc Bunny} & AVG \\
    \hline
    \multirow{2}[0]{*}{{\sc Nylon}} 
          & DANI-Net & 7.06  & 20.98 & 22.63 & 22.79 & 25.33 & 28.3  & 26.96 & 24.00 & \textbf{34.37} & 23.82 & 23.62 \\
          & DANI-Net {\it w rm}& \textbf{5.23} & \textbf{13.40} & \textbf{9.61} & \textbf{16.22} & \textbf{20.43} & \textbf{24.10} & \textbf{19.18} & \textbf{22.62} & 34.43  & \textbf{19.17} & \textbf{18.44} \\
    \hline
    \multirow{2}[0]{*}{{\sc Acrylic}}
          & DANI-Net & \textbf{34.14} & 39.13 & 36.63 & 35.25 & 42.70  & 36.90  & \textbf{44.67} & \textbf{23.50}  & 33.03 & 21.89 & 34.78  \\
          & DANI-Net {\it w rm} & 35.27 & \textbf{25.50} & \textbf{35.10} & \textbf{33.38} & \textbf{36.98}  & \textbf{36.16}  & 45.61 & 25.23  & \textbf{32.90}  &  \textbf{19.27} & \textbf{32.54} \\
    \hline
    \end{tabular}%
    }
\end{table*}

\begin{figure}[h]
    \centering
    \includegraphics[width=0.85\textwidth]{"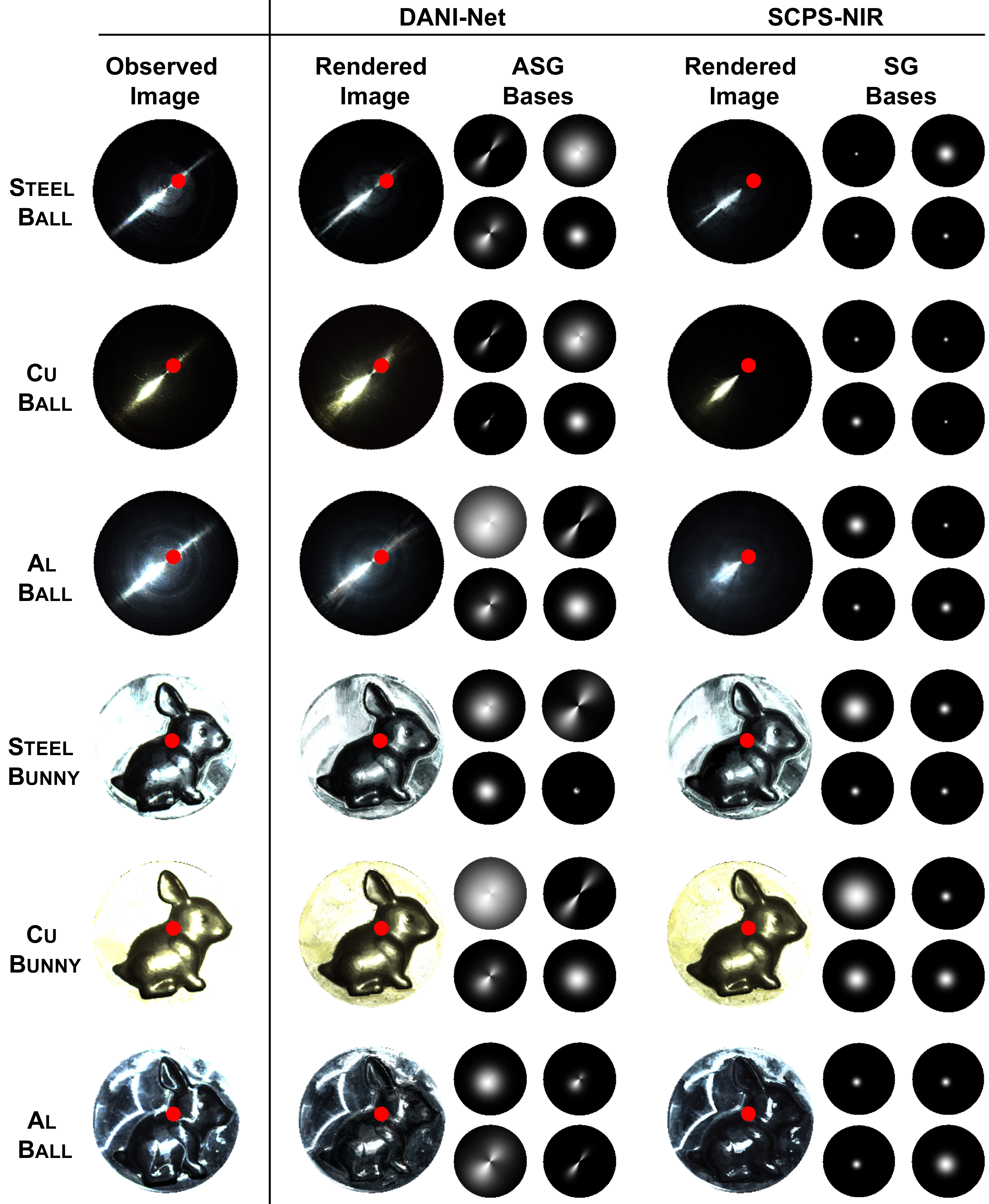"}
    \caption{Visualization of the observed images (column 1), rendered images (column 2, 5), ASG bases (column 3, 4) and SG bases (column 6, 7) on {\sc Ball} and {\sc Bunny} made up of {\sc Steel}, {\sc Cu}, and {\sc Al}. The rendered images with anisotropic material modeling are generated by DANI-Net (column 2-4). The rendered images with isotropic material modeling are generated by SCPS-NIR~\cite{li2022self} (column 5-7).}
    \label{fig:mat}
\end{figure}

\clearpage
\newpage
\textbf{Validation of anisotropic material modeling.}
To validate the effectiveness of our anisotropic material modeling method, we visualize the rendered image, and 4 dominant ASG bases of 2 shapes (\ie, {\sc Ball} and {\sc Bunny}) made up of 3 anisotropic materials (\ie, {\sc Steel}, {\sc Cu}, and {\sc Al} respectively), compared with SCPS-NIR~\cite{li2022self}'s rendered image and 4 dominant SG bases in \Fref{fig:mat}. According to the result, our rendered images are more realistic than~\cite{li2022self}.

\clearpage
\newpage
\section{Results on {\sc Apple \& Gourd} Dataset~\cite{alldrin2008photometric} and {\sc Light Stage Data Gallery} dataset~\cite{chabert2006relighting}}
\label{sec:comp_apple_light}
\textbf{Preprocessing.} For {\sc Apple \& Gourd}, we scale the images into 0.5 of their original size for efficient training. For {\sc Light Stage Data Gallery}, we use a preprocessing method that is different from~\cite{li2022self, chen2020learned}. We follow~\cite{chabert2006relighting} to implement gamma correction with an exponent as $2.2$ on objects with LDR observed images 
(\ie objects expect {\sc Helmet Front})
and magnify the pixel intensity 5 times on objects except {\sc Plant} because the original pixel intensities on those objects are too low, leading to strong artifacts on the predicted normal map. We also downsample the {\sc Plant}'s observed images to 0.5 of their original size for efficient training. The above preprocessing process is also applied in~\cite{li2022self} for a fair comparison.

\begin{figure}[h]
    \centering
    \includegraphics[width=0.8\textwidth]{"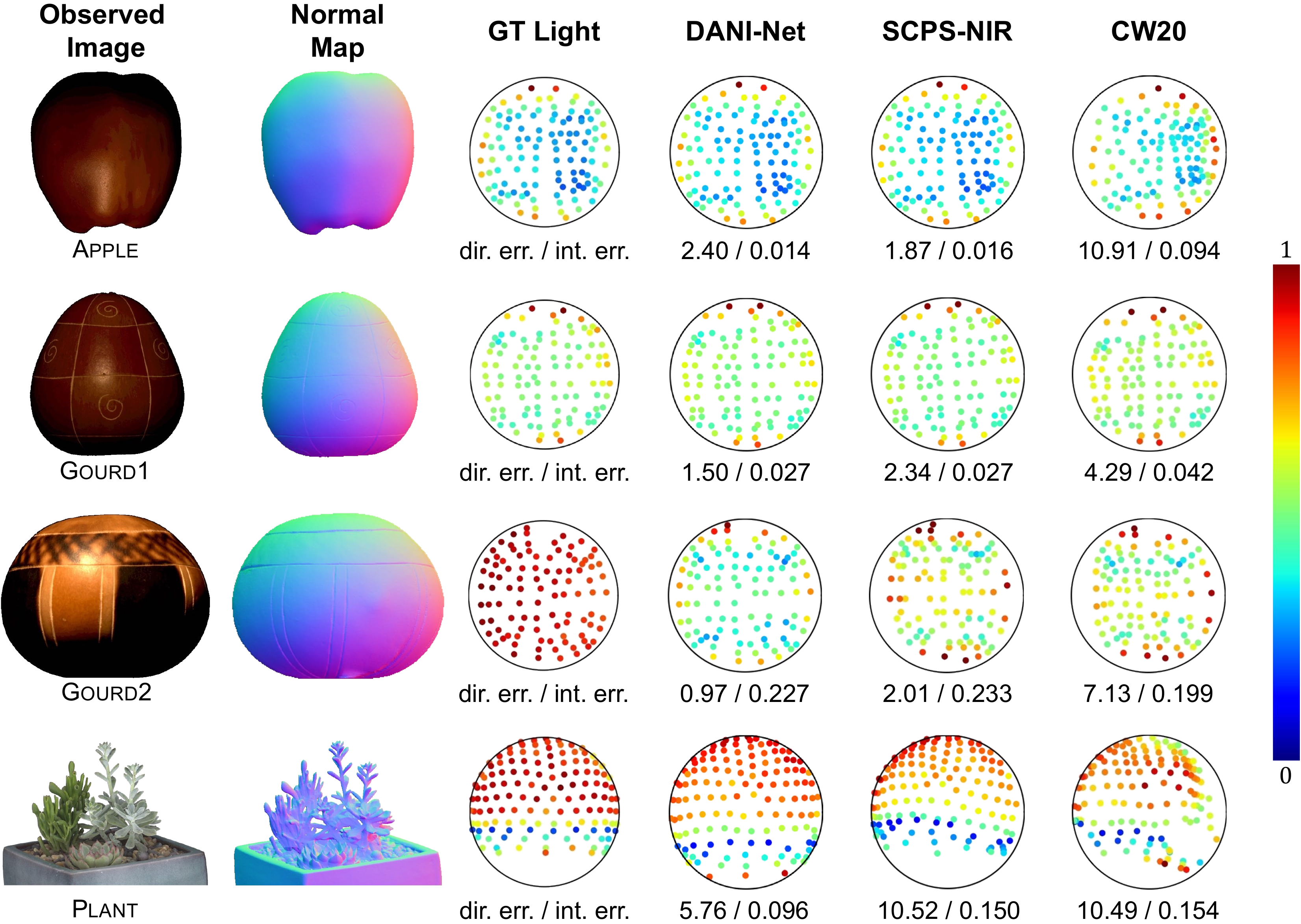"}
    \caption{The visual quality comparison among DANI-Net, SCPS-NIR~\cite{li2022self}, and CW20~\cite{chen2020learned} on {\sc Apple}, {\sc Gourd1}, and {\sc Gourd2} from {\sc Apple \& Gourd}~\cite{alldrin2008photometric} and {\sc Plant} from {\sc Light Stage Data Gallery}~\cite{chabert2006relighting} in terms of normal map (column 2) and light map (column 3 - 6). Numbers indicate the MAE (for light directions).}
    \label{fig:gallery}
\end{figure}

\begin{figure}[h]
    \centering
    \includegraphics[width=0.9\textwidth]{"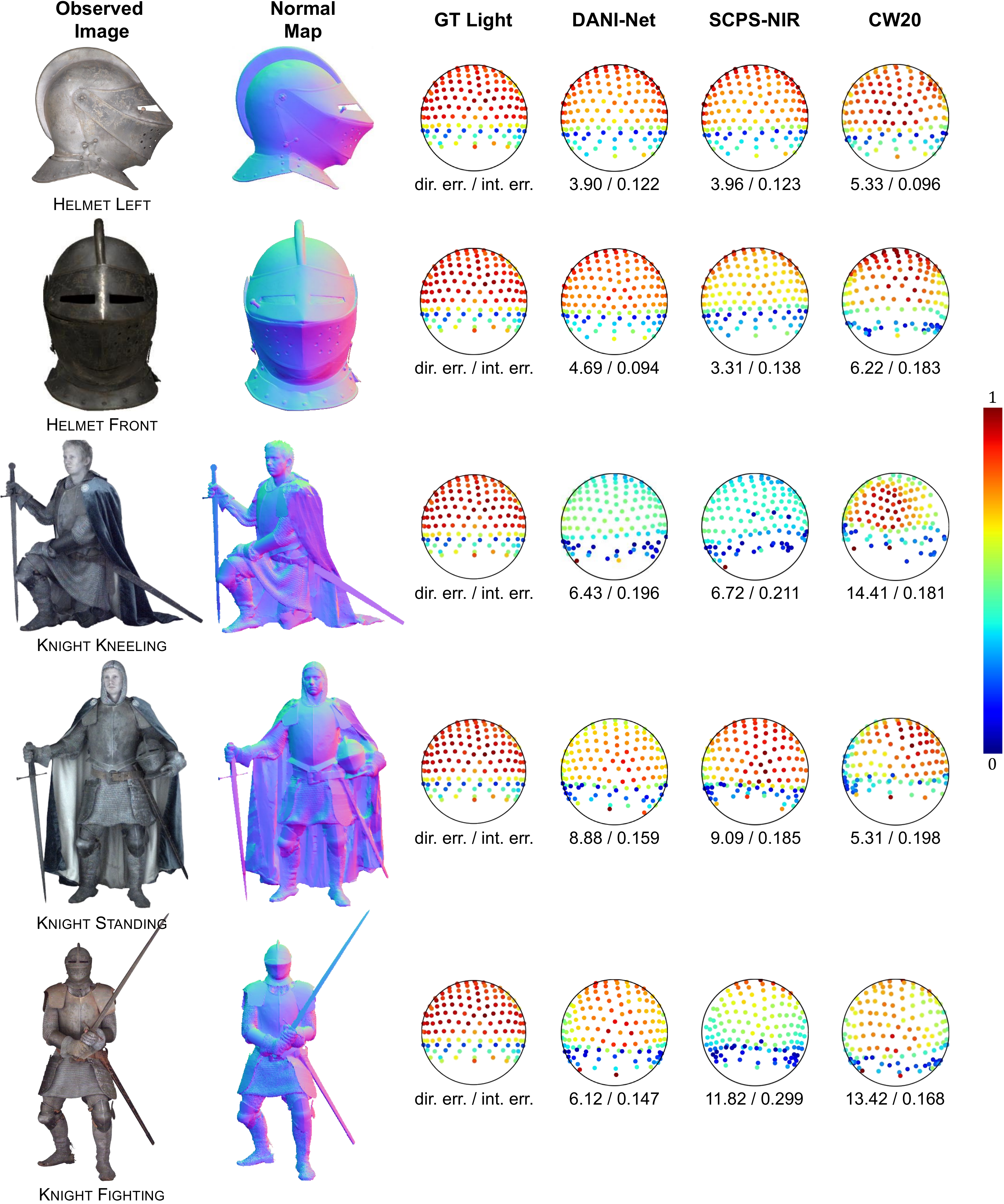"}
    \caption{The visual quality comparison among DANI-Net, SCPS-NIR~\cite{li2022self}, and CW20~\cite{chen2020learned} on {\sc Helmet Left}, {\sc Helmet Front}, {\sc Knight Kneeling}, {\sc Knight Standing} and {\sc Knight Fighting} from {\sc Light Stage Data Gallery}~\cite{chabert2006relighting} in terms of normal map (column 2) and light map (column 3 - 6). Numbers indicate the MAE (for light directions).}
    \label{fig:gallery2}
\end{figure}

\textbf{Qualitative and Quantitative Results.} \Tref{tab:light} shows the light calibration results of 3 objects from {\sc Apple \& Gourd} Dataset~\cite{alldrin2008photometric} and 6 objects from {\sc Light Stage Data Gallery} dataset~\cite{chabert2006relighting}. As these two datasets don't release the ground truth normal map, we only provide the predicted normal map in \Fref{fig:gallery} and \Fref{fig:gallery2} for qualitative analysis. The results show that DANI-Net can not only estimate the surface normal reasonably but also calibrate the light accurately (we have the state-of-the-art light calibration results), illustrating that DANI-Net is free from the deviation of different experimental setups. 

\begin{table*}[h]
    \centering
    \setlength{\tabcolsep}{9pt}
    \setlength{\abovecaptionskip}{0cm}
    \setlength{\belowcaptionskip}{0cm}
    \caption{Quantitative comparison in terms of MAE of light direction and scale-invariant error of intensity on {\sc DiLiGenT} benchmark dataset~\cite{shi2016benchmark}. \textbf{Bold numbers} and \underline{underlined numbers} indicate the best and the second-best results, respectively. }
    \label{tab:light}
    \resizebox{\linewidth}{!}{
    \begin{NiceTabular}{c|cccccc|cccccccccccc|cc}
    \hline
    \multirow{2}[2]{*}{Model}& \multicolumn{2}{c}{\sc{Apple}} & \multicolumn{2}{c}{\sc{Gourd1}} & \multicolumn{2}{c}{\sc{Gourd2}} & \multicolumn{2}{c}{\sc Plant} & \multicolumn{2}{c}{\makecell{\sc{Helmet} \\ \sc{Left}}} &  \multicolumn{2}{c}{\sc{\makecell{Helmet \\ Front}}} &  \multicolumn{2}{c}{\sc{\makecell{Knight \\ Kneeling}}} & \multicolumn{2}{c}{\sc{\makecell{Knight \\ Standing}}} & 
    \multicolumn{2}{c}{\sc{\makecell{Knight \\Fighting}}} &
    \multicolumn{2}{c}{AVG} \\ 
    & dir. & int.  & dir. & int.  & dir. & int.  & dir. & int. & dir. & int. & dir. & int. & dir. & int. & dir. & int. & dir. & int. & dir. & int. \\
    \hline
    PF14~\cite{papadhimitri2014closed} & 6.68 & 0.109 & 21.23 & 0.096 & 25.87 & 0.329 & 20.56 & 0.227 & 25.40 & 0.576 & 81.60 & \underline{0.133} & 46.69 & 9.805 & 33.81 & 1.311 & 69.50 & 1.137 & 36.82 & 1.52 \\
    CW20~\cite{chen2020learned} & 10.91 & 0.094 & 4.29 & 0.042 & 7.13 & \textbf{0.199} & \underline{10.49} & 0.154 & 5.33 & \textbf{0.096} & 6.22 & 0.183 & 14.41 & \textbf{0.181} & \textbf{5.31} & 0.198 & 13.42 & \underline{0.168} & 8.61 & 0.146\\
    SCPS-NIR~\cite{li2022self} & \textbf{1.87} & \underline{0.016} & \underline{2.34} & \textbf{0.027} & \underline{2.01} & \underline{0.233} & 10.52 & \underline{0.150} & \underline{3.96} & 0.123 & \textbf{3.31} & 0.138 & \underline{6.72} & 0.211 & 9.09 & \underline{0.185} & \underline{11.82} & 0.299 & \underline{5.74} & \underline{0.154}\\
    \hline
    DANI-Net & \underline{2.40}  & \textbf{0.014}  & \textbf{1.50}  & \textbf{0.027}  & \textbf{0.97}  & 0.227 & \textbf{5.76} & \textbf{0.096} & \textbf{3.90} & \underline{0.122}  & \underline{4.69}  & \textbf{0.094} & \textbf{6.43} & \underline{0.196}  & \underline{8.88}  & \textbf{0.159}  & \textbf{6.12} & \textbf{0.147} & \textbf{4.74} & \textbf{0.129}\\
    \hline
    \end{NiceTabular}
    }
\vspace{-5pt}
\end{table*}

\clearpage
\newpage
\end{document}